\documentclass{article}
\usepackage[utf8]{inputenc}
\usepackage{times,latexsym}
\usepackage{url}
\usepackage[T1]{fontenc}
\usepackage{times}
\usepackage{latexsym}
\usepackage{amsmath}
\usepackage{amssymb}
\usepackage{wasysym}
\usepackage{bbm}

\usepackage{booktabs}
\usepackage{multirow}

\usepackage{bibentry}
\nobibliography*

\usepackage{color,soul,colortbl}

\title{Robust Conversational AI with Grounded Text Generation}

\author{
Jianfeng Gao, Baolin Peng, Chunyuan Li, Jinchao Li \\
Shahin Shayandeh, Lars Liden, Heung-Yeung Shum\\
\\
Microsoft Research \\
\small{\{jfgao, bapeng, chunyl, jincli, shahins, laliden, v-hshum\}@microsoft.com}
}

\date{September 7, 2020}

\usepackage{natbib}
\usepackage{graphicx}

\begin{document}
\maketitle

\begin{abstract}
This article presents a hybrid approach based on a Grounded Text Generation (GTG) model to building robust task bots at scale.  GTG is a hybrid model which uses a large-scale Transformer neural network as its backbone, combined with symbol-manipulation modules for knowledge base inference and prior knowledge encoding, to generate responses grounded in dialog belief state and real-world knowledge for task completion.
GTG is pre-trained on large amounts of raw text and human conversational data, and can be fine-tuned to complete a wide range of tasks.  

The hybrid approach and its variants are being developed simultaneously by multiple research teams. The primary results reported on task-oriented dialog benchmarks are very promising, demonstrating the big potential of this approach. This article provides an overview of this progress and discusses related methods and technologies that can be incorporated for building robust conversational AI systems.



\end{abstract}

\section{Introduction}
\label{sec:introduction}
The long-term mission of conversational AI research is to develop at scale conversational assistant systems, also known as \emph{task-oriented bots} or \emph{task bots} in short, 
which are \emph{robust} enough that 
(1) they can help users accomplish various tasks ranging from question answering and restaurant reservation to travel planning,
(2) their responses are always interpretable, controllable, and reliable, even in a highly dynamic environment (e.g., due to users changing back and forth among different tasks and topics), and 
(3) they can transfer the knowledge and skills learned in one task to other tasks. 

Despite decades of research, the mission remains unfulfilled. 
Almost all task bots used in real-world applications are developed using task-specific, hand-crafted rules and programs -- an approach that  fundamentally does not scale.  
Although machine learning methods are critical to the development of many robust NLP systems, such as machine translation and speech recognition, they play a far less important role in building task bots. 
For example, deep-learning based neural approaches to conversational AI, which become increasingly important as a research area \citep{gao2019neural}, have not widely used for building commercial task bots yet because they are not robust enough.

\begin{figure}[t] 
\centering 
\includegraphics[width=0.95\linewidth]{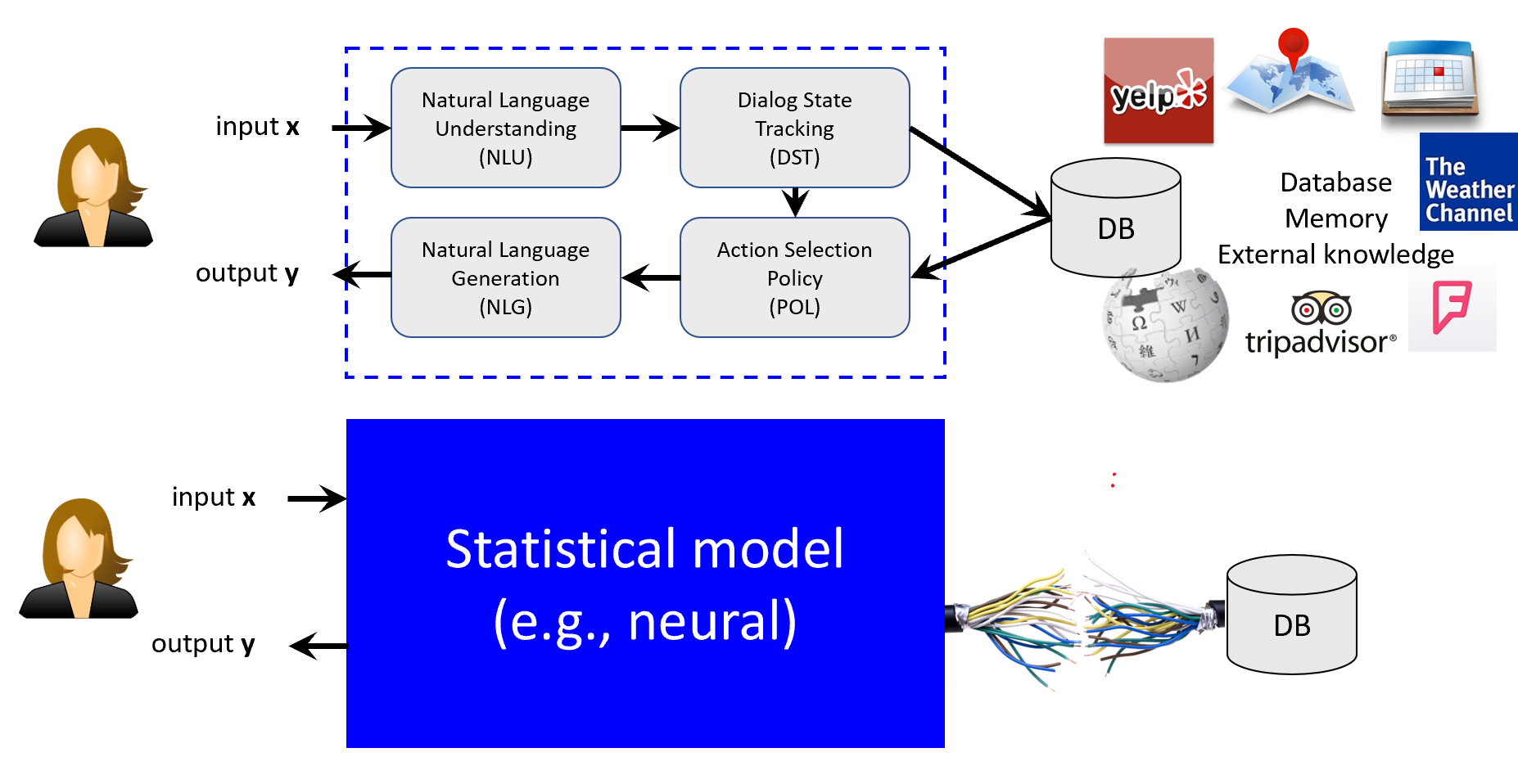}
\vspace{-2mm}
\caption{Two architectures of dialogue systems \cite{gao2019neural}: (Top) classical task-oriented dialog and (Bottom) fully data-driven end-to-end dialog. } 
\label{fig:two-dialogue-system} 
\vspace{0mm}
\end{figure}

Since 2019, we have witnessed a paradigm shift in conversational AI research due to the progress on large-scale pre-trained language models for text generation such as the Generative Pre-Training (GPT) models \cite{radford2018improving,gpt2,gpt3}. Instead of using the classical modular architecture of task bots, as shown in Figure \ref{fig:two-dialogue-system}(Top), which is composed of multiple modules for natural language understanding, dialog state tracking, action selection, and response generation, respectively, some of the latest chatbots, which are designed primarily for chitchat rather than task completion, are developed using a unitary system, as shown in Figure \ref{fig:two-dialogue-system}(Bottom), which directly generates natural language response given user input using a neural language model (such as GPT-2) trained on large amounts of human-conversational data. Although the neural language model is not designed for task completion, it can generate fluent and human-like responses to any user input \cite{gao2019neural,zhang2019dialogpt,adiwardana2020towards,gpt2,gpt3}.  Thus, there are discussions on adopting the same unitary system to build task bots at scale, based on the assumption that by simply increasing the amount of training data and model capacities, these chatbots will be able to learn to complete all sorts of tasks and evolve themselves into task bots.

In the first half of this paper, we show that such an assumption, despite its optimism, is not grounded due to the fundamental limitation of GPT and other similar large-scale language models such as BERT \cite{devlin2019bert}, XLNet \cite{yang2019xlnet}, UniLM \cite{dong2019unified}. These large-scale neural language models are designed to learn language patterns (i.e., how words co-occur with one another in large text corpora) for text prediction, whereas task bots need to detect user intents and take a sequence of goal-directed actions, grounded in task-specific knowledge and dialog belief state, for task completion. 
We will show that the language models are by no means sufficient to building robust task bots, no matter how large the training data and models we use. Although the generated responses are fluent and plausible-sounding, users cannot count on them to complete any specific tasks since they are not explicitly grounded in user intents and task-specific, real-world knowledge bases.

The classical modular architecture of task bots, on the other hand, is grounded in the theories of human cognition and communication. 
But the classical approach has proved not scalable since it requires task-specific labels to train individual dialog modules for each task, and such task labels are not available in large amounts for many real-world scenarios.

In the second half of the paper, we present a hybrid approach based on a Grounded Text Generation (GTG) model to building robust task bots at scale, combining the best of both large-scale language model pre-training and the classical modular architecture of task bots.
The hybrid approach and its variants are being developed simultaneously by multiple research teams \cite{peng2020soloist,hosseini2020simple,Ham2020e2e,budzianowski2019hello,zhang2019task,lei2018sequicity}. The results reported so far on research benchmarks demonstrate the big potential of the approach. We will provide an overview of this progress and discuss related methods and technologies that can be incorporated for building robust task bots. 

In the hybrid approach, we follow the classical modular architecture to develop task bots that are equipped with cognitive models to track dialog belief state and take goal-directed actions to complete tasks. Unlike classical modular task bots, our implementation leverages state-of-the-art deep learning technologies for scalable modeling. Our task bot is a multi-turn decision-making system based on a pre-trained GTG – a hybrid model that uses a large-scale Transformer neural network as its backbone, combined with symbol-manipulation modules for knowledge base inference and encoding commonsense knowledge and business rules for action selection and natural language response generation. Specially, we view a modular system as a sequential data generation pipeline where the output of a module is the input of the next module. We implement each module as a Transformer neural network and allow all these neural networks to share their parameters. Then, the whole pipeline is implemented using a Transformer model. While GPT-2 is an autoregressive language for text prediction, GTG is a stateful model for decision making.

While the classical approach to building task bots requires training each module for each task individually using task labels, the GTG-based hybrid approach leverages the pre-training and fine-tuning paradigm to build robust task bots at scale. Specifically, GTG is pre-trained and fine-tuned for the completion of individual tasks in three stages.
\begin{enumerate}
    \item {\bf Language pre-training} via self-supervised learning. The GTG model learns the primary skills of understanding and generating natural language on large amounts of raw text. The pre-trained model encodes text as a sequence of symbolic tokens without grounding them in real-world concepts \cite{lucy2017distributional} and generates text by searching word co-occurrence void of meaning \cite{bisk2020experience}. Therefore, the model might generate fluent responses that are not useful to achieve any specific goals.
    \item {\bf Task completion pre-training} via supervised learning. GTG learns the primary skills of task completion, including tracking dialog belief states and user intents to build internal cognition models, knowledge base lookup (via neuro-symbolic reasoning), and deciding how to respond to complete a task. The pre-trained model grounds language words to perceptions and real-world concepts, and generates responses in terms of causality. 
    \item {\bf Task-specific fine-tuning} via supervised learning and reinforcement learning. GTG learns to adapt itself, in a fully embodied and social context, to complete specific tasks using a hybrid learning framework based on reinforcement learning and machine teaching, which is a form of supervised learning where training samples are generated by human teachers interacting with the system, as detailed in Section \ref{subsubsec:supervised-learning}, so that (1) the responses generated by GTG are grounded in task-specific rules and knowledge base that are encoded into the system via coding and machine learning, (2) the responses are optimized for task completion, and (3) the primary skills are constantly improved task by task so that the bot can adapt to new tasks more easily. 
\end{enumerate}


The rest of the paper is structured as follows. 
Section 2 discusses the fundamental limitations of using neural language models for task completion. 
Section 3 revisits the classical modular approach to task bots and reviews early attempts to make the approach scalable by incorporating neural methods. 
Section 4 describes the GTG-based hybrid approach to building robust task bots at scale, and provides an overview of this progress in the community.
Section 5 summarizes the results reported on research benchmarks to demonstrate the potential of the hybrid approach.
Section 6 concludes the paper by discussing the challenges we are facing and their possible solutions.

It is worth noting that our discussion in this paper is also inspired by Gary Marcus's critical comments on deep learning \cite{marcus2020next,marcus2019rebooting,marcus2018deep}. 
In a sense, the GTG-based approach to robust conversational AI can be viewed as an instance of the hybrid approach to robust AI advocated in \cite{marcus2020next}.

\section{Neural Language Models for Task Bots}
\label{sec:neural-language-model-bot}


Since 2015, researchers have begun to explore a fully data-driven and end-to-end approach based on neural language models to social chatbots, where a multi-turn dialog session is modeled as a long text sequence and the response generation task as language modeling \cite{shang2015neural,sordoni2015neural,vinyals2015neural,li2016diversity,serban2016generative,gao2019neural,roller2020open}.

\begin{figure}[t] 
\centering 
\includegraphics[width=0.7\linewidth]{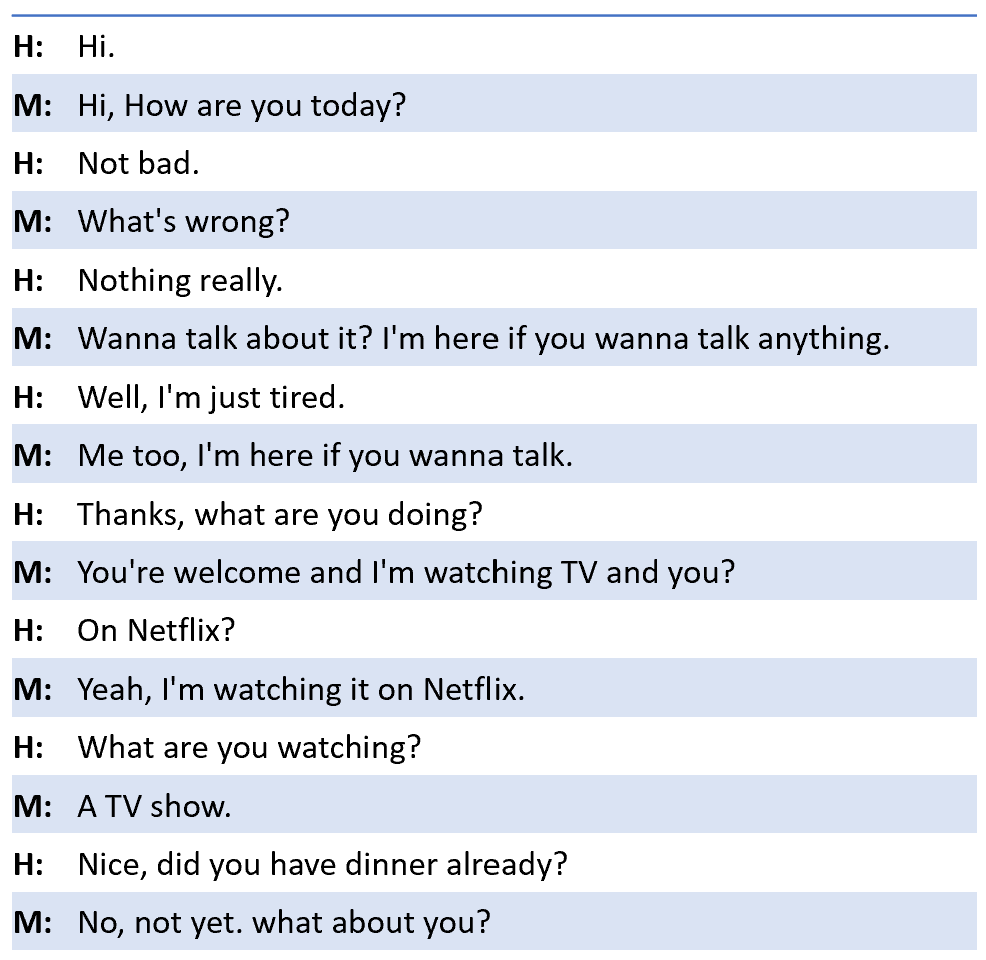}
\vspace{-1mm}
\caption{A dialog session generated by a human user (H) and a neural language model based chatbot system (M). The bot can generate responses that are fluent and remarkably on topic.} 
\label{fig:chatbot-output} 
\vspace{0mm}
\end{figure}

This approach is appealing for two reasons. First, these language models are trained entirely on data without resorting to any expert knowledge or additional handcrafting. Second, it easily scales to large free-form and open-domain datasets, which means the trained chatbots can typically chat on any topic. 
Recent papers \cite{zhang2019dialogpt,adiwardana2020towards,roller2020recipes} show that with an increase of the model capacity and the amount of human conversational data for training, these chatbots demonstrate more human-like intelligence, and can generate responses that are so fluent and remarkably on topic that they are indistinguishable from human-generated ones, as shown in Figure~\ref{fig:chatbot-output}.
Thus, there is a growing interest in adopting these chatbots for task completion \cite{budzianowski2019hello}.

However, the language model based approach suffers from three fundamental problems that need to be addressed before it is adequate for building robust task bots. 
These language model based chatbots cannot reliably generate responses grounded in real-world knowledge, do not have internal cognitive models to keep track of dialog states, and cannot generate goal-directed responses that are optimized for task completion. 
Below, we elaborate these problems using examples.  

\begin{figure}[t] 
\centering 
\includegraphics[width=0.7\linewidth]{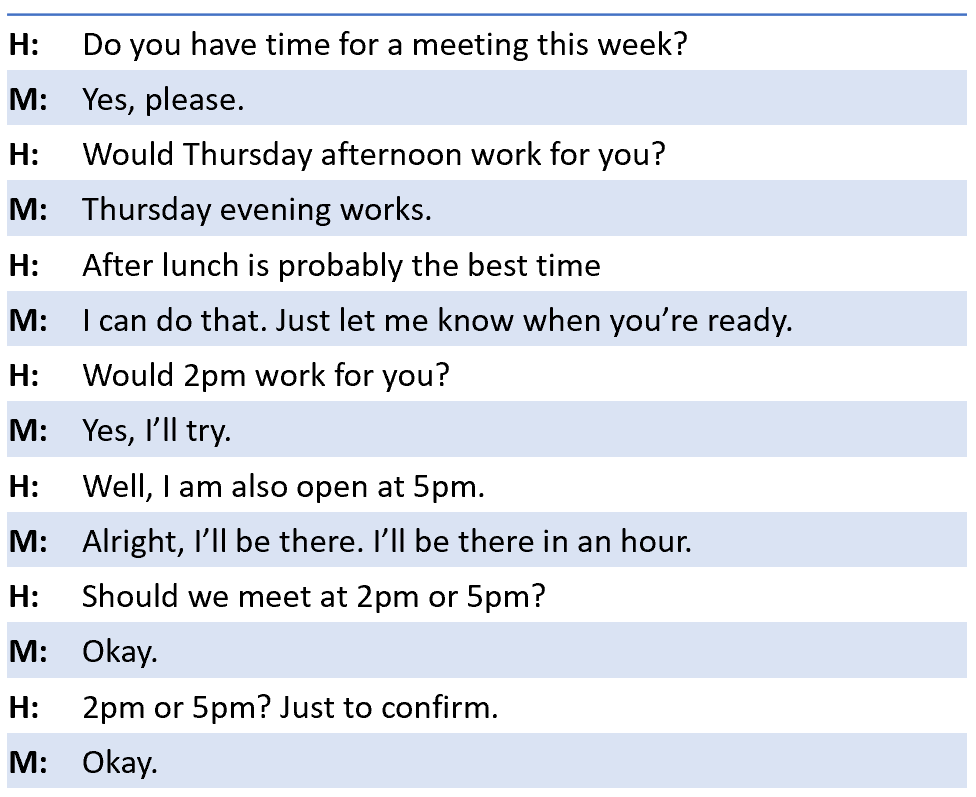}
\vspace{-1mm}
\caption{A task-oriented dialog session (scheduling a meeting) generated by a human user (H) and a neural language model based chatbot system (M). 
The bot's responses are unreliable since they are not grounded in any real calendar but generated based on word distribution learned from training data. 
The bot also fails to keep track of the meeting time suggested by the user and cannot give a proper confirmation.} 
\label{fig:taskdialog-by-chatbot} 
\vspace{0mm}
\end{figure}

\begin{figure}[t] 
\centering 
\includegraphics[width=0.7\linewidth]{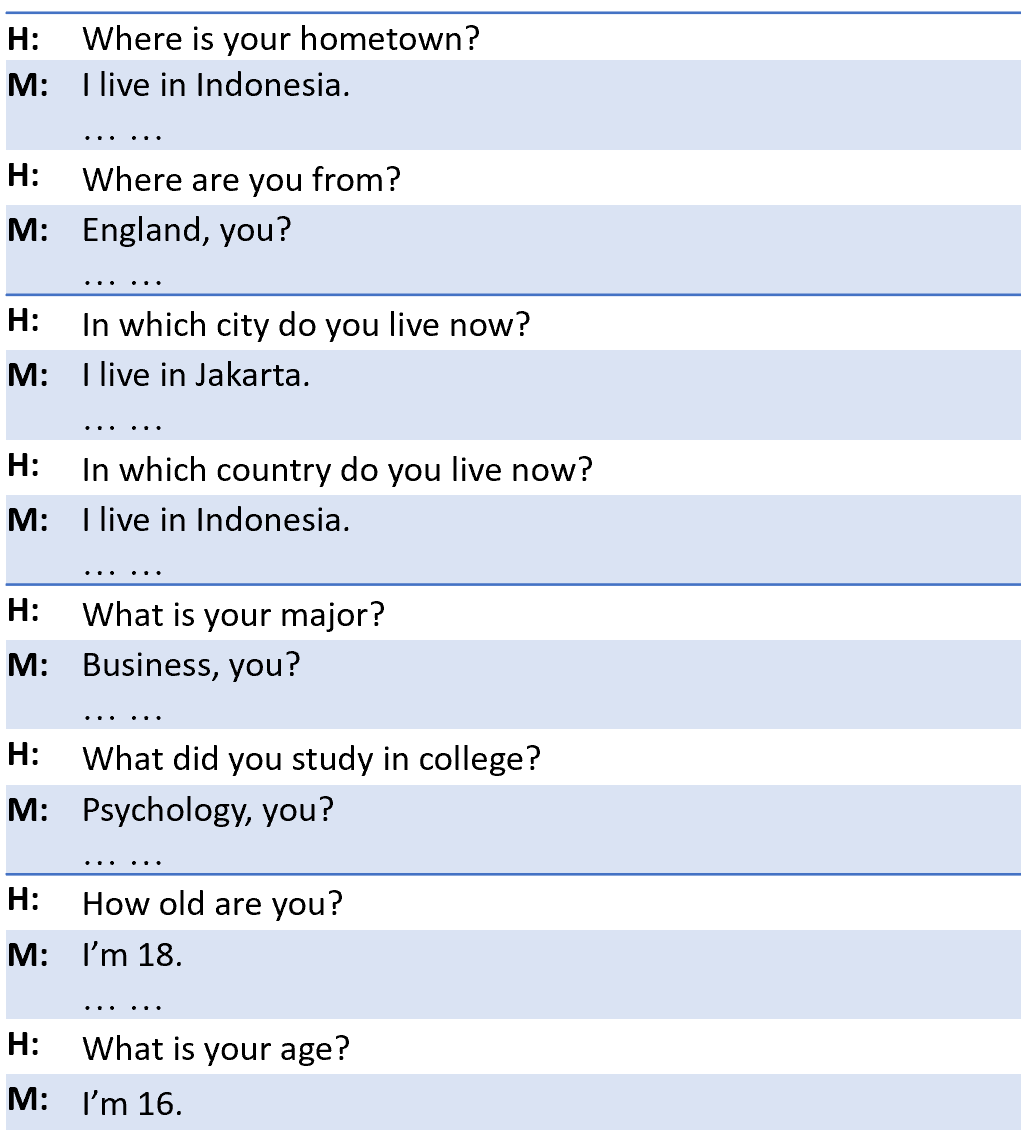}
\vspace{-1mm}
\caption{Inconsistent responses generated by a neural language model based chatbot that is not grounded in any persona model. Adapted from \cite{li2016persona}.} 
\label{fig:ungrounded-persona-output} 
\vspace{0mm}
\end{figure}

Figure~\ref{fig:taskdialog-by-chatbot} shows a task-oriented dialog between a human user and a language model based chatbot \cite{sordoni2015neural,li2016diversity}. 
The conversation is as fluent and plausible as the chitchat dialog in Figure~\ref{fig:chatbot-output}. 
In addition, the bot seems to understand the user's intent and agrees to meet the user at 2pm. But this is an illusion! If the user went to the meeting room as scheduled, no one would have shown up. 
The bot only learns the \emph{shape} of the conversation without a deep understanding of the user's intent. The bot's responses are unreliable since they are not grounded in any real calendar but generated based on the word distribution learned from the training data. 
The lack-of-grounding issue has also been demonstrated in the consistency test for evaluating personal social chatbots \cite{li2016persona}. 
As shown in the examples in Figure~\ref{fig:ungrounded-persona-output}, a chatbot that is not grounded in a persona model (or a user profile) gives inconsistent responses even within a dialog session. 
The problem remains after we enrich the language model to capture longer distance dependencies, trained on much larger amounts of conversational data, as in DialoGPT \cite{zhang2019dialogpt}. 

\begin{figure}[t] 
\centering 
\includegraphics[width=0.99\linewidth]{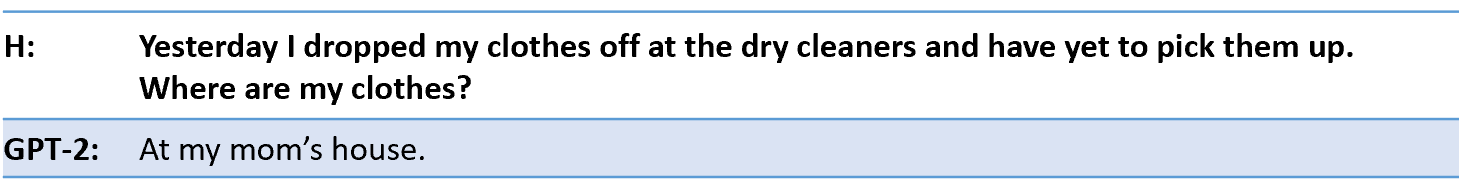}
\vspace{-1mm}
\caption{Given the user input and question (in \textbf{bold}), GPT-2 generates a wrong answer since it cannot keep track of the location of the clothes. Adapted from \cite{marcus2020next}.} 
\label{fig:gpt2-output} 
\vspace{0mm}
\end{figure}

In the example in Figure~\ref{fig:taskdialog-by-chatbot}, after the user suggests meeting at 5pm instead of 2pm, the bot, which is not equipped with an internal cognitive model to keep track of the dialog state, fails to update the meeting time suggested by the user and cannot give a proper confirmation. A similar example is given by Marcus \cite{marcus2020next}, as in Figure~\ref{fig:gpt2-output}, where the GPT-2 model also fails to respond correctly because the model cannot keep track of ``where my clothes are''.

\begin{figure}[t] 
\centering 
\includegraphics[width=0.99\linewidth]{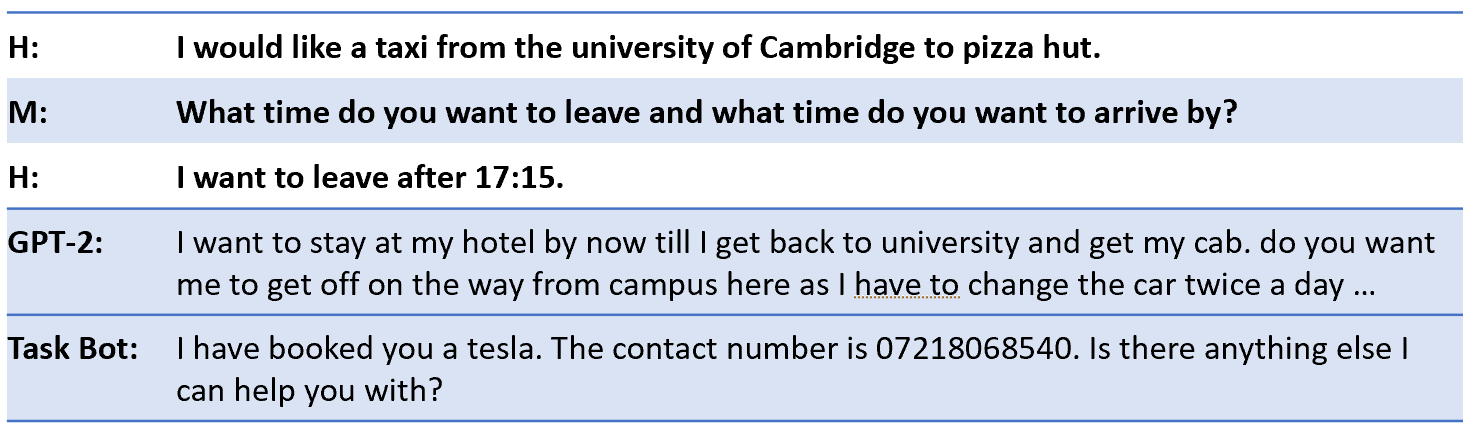}
\vspace{-1mm}
\caption{Given a dialog history (in \textbf{bold}), the response generated by GPT-2 is fluent and plausible but is not useful for completing the task whereas the response generated by a Task Bot is concise and to the point.} 
\label{fig:task-bot-vs-gpt2} 
\vspace{0mm}
\end{figure}

Figure~\ref{fig:task-bot-vs-gpt2} shows the responses generated by GPT-2 and a task bot \cite{peng2020soloist}, respectively. We can see that given the dialog history, although GPT-2 can produce very fluent and plausible-sounding continuation that captures, for example, the relation between locations (hotel, university, campus), cab and time, it does not provide any useful information (e.g., the contact number of the taxi) to complete the task. In contrast, the task bot's response 
is concise and to the point.  

The aforementioned three problems cannot be addressed by simply increasing the amount of training data and model capacity. 
They are rooted in the inherent limitation of deep learning models being used as black boxes and thus not interpretable. 
The responses of a task bot, on the other hand, need to be not only 
interpretable but controllable so that their designers always know why the bot responds in a particular way and, if necessary, are able to specify using rules or task-specific codes to control how the bot responds in a given context. 

The solution we advocate in this paper is a hybrid approach that uses deep learning to build, at scale, robust task bots that can generate goal-directed responses explicitly based on task-specific knowledge and cognitive models. 
Before we present the hybrid approach in detail in Section~\ref{sec:gtg}, in the next section we revisit the classical approach to task bots where the primary focuses are on knowledge, dialog state tracking, and goal-directed action selection.   



\section{Revisit of Classical Modular Task Bots}
\label{sec:modular-bot}

The classical modular approach to building task bots is motivated by the theories of human cognition. 
Cognition is formulated as an iterative decision making process \cite{marcus2020next}: organisms (e.g., humans) take in information from the environment, build internal cognitive models based on their perception of that information, which includes information about the entities in the external world, their properties and relationships, and then make decisions with respect to these cognitive models which lead to human actions that change the environment. Cognitive scientists generally agree that the degree to which an organism prospers in the world depends on how good those internal cognitive models are \cite{gallistel1990organization,gallistel2011memory}. 

Similarly, the classical modular architecture of task bots, as shown in Figure~\ref{fig:two-dialogue-system}(Top), views multi-turn conversations between a bot and a user as an iterative decision making process, where the bot is (the agent of) the organism and the user the environment. 
The bot consists of a pipeline of modules that play different roles in decision making. At each iteration, a natural language understanding (NLU) module identifies the user intent and extracts associated information such as entities and their values from user input. A dialog state tracker (DST) infers the dialog belief state (the internal cognitive model of the bot). The belief state is often used to query a task-specific database (DB) to obtain the DB state, such as the number of entities that match the user goal. The dialog state and DB state are then passed to a dialog policy (POL) to select the next system action. A natural language generation (NLG) module converts the action to a natural language response.
Like cognitive scientists, dialog researchers also believe that the quality of task bots depends to a large degree upon the performance of dialog state tracking (or their internal cognitive models), which had been the focus of task-oriented dialog research for many years \cite{williams13dialog,henderson14second,henderson14third,kim16fourth,kim16fifth,hori17sixth}.

Different from the language model based approach described in Section~\ref{sec:neural-language-model-bot}, the modular approach allows to build task bots whose behaviors are interpretable, controllable, and goal-directed. 
Thus, almost all commercial tools for building task bots employ the modular approach, including 
Google's Dialog Flow\footnote{{https://dialogflow.com/}},
Microsoft's Power Virtual Agents (PVA)\footnote{{https://powervirtualagents.microsoft.com/}}, 
Facebook's Wit.ai\footnote{{https://wit.ai/}}, 
Amazon's Lex\footnote{{https://aws.amazon.com/lex/}}, and 
IBM's Watson Assistant\footnote{{https://www.ibm.com/watson/}}.

\begin{figure}[t] 
\centering 
\includegraphics[width=0.95\linewidth]{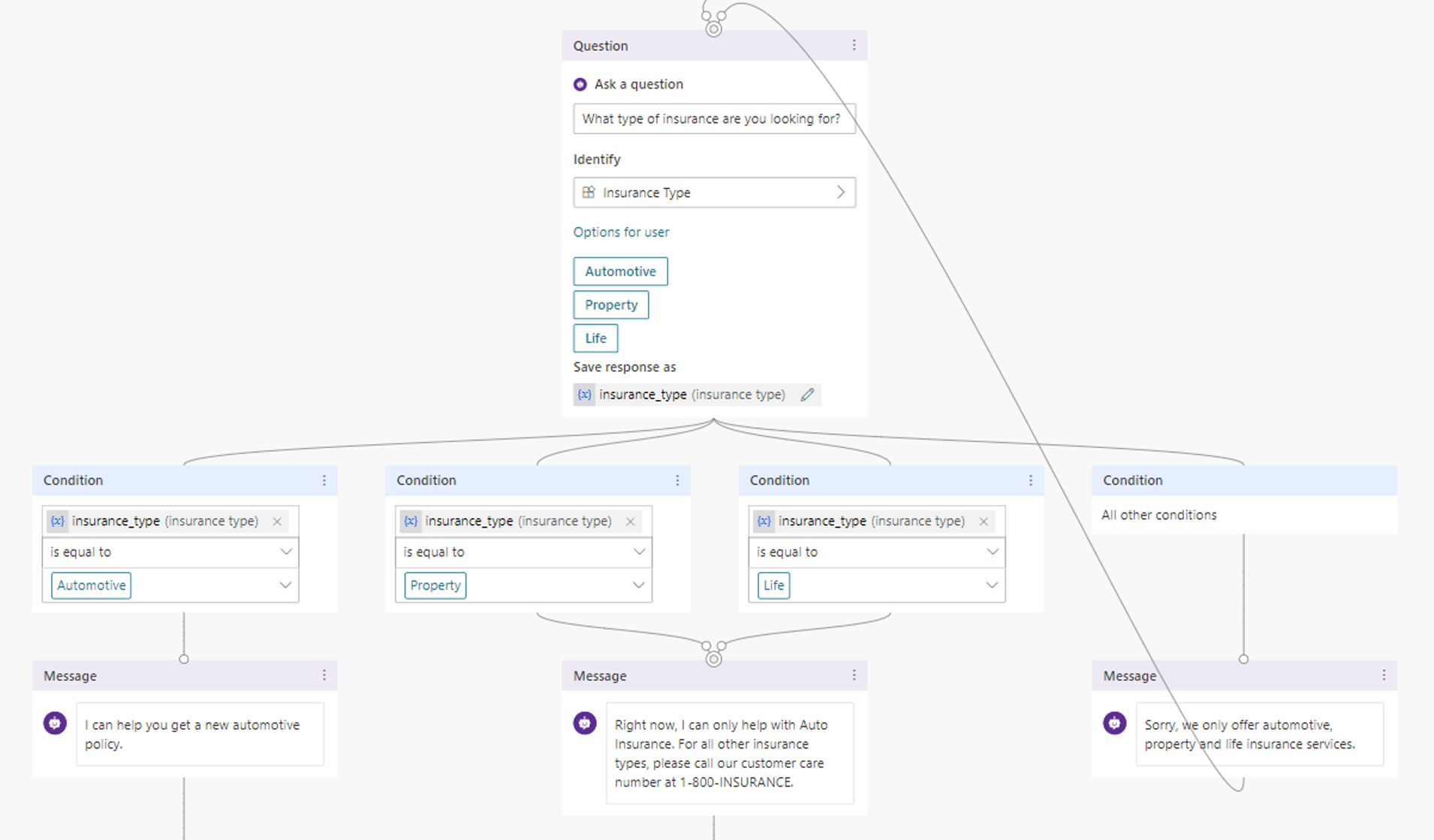}
\vspace{-1mm}
\caption{An example of a dialog flow specification using Microsoft's Power Virtual Agents.} 
\label{fig:pva} 
\vspace{0mm}
\end{figure}

These tools are known as dialog composers. 
They are often implemented as drag-and-drop WYSIWYG tools that allow dialog authors to specify and visualize all the details of the dialog flow. 
They often have deep integration with popular Integrated Development Environments (IDEs) as editing front-ends.
For example, Microsoft's PVA expresses a dialog flow as a finite-state machine, with nodes representing dialog actions and arcs corresponding to states. Figure~\ref{fig:pva} illustrates an example of a graphical dialog flow specification, where dialog authors need to explicitly specify dialog states (e.g., conditions), and for each state system actions (e.g., messages).  
However, PVA can only handle simple dialog tasks where the number of dialog states and actions is limited. The dialog flow can grow quickly to be too complex to manage as the task complexity increases or `off-track'' dialog paths have to be dealt with to improve the robustness of task bots. 

Building task bots at scale by leveraging machine learning (ML) and large amounts of dialog data has been the focus of the research community for decades.   
Over the years, different ML dialog models have been developed for (1) individual dialog modules (e.g., NLU, DST, POL, NLG, and DB lookup, as in Figure~\ref{fig:two-dialogue-system}(Top)) (2) end-to-end task-oriented dialog systems \cite{li2017end,lei2018sequicity,zhang2019task}, or (3) jointly modeling some of the typical dialog modules. 
For example, the word-level state tracking models \cite{ramadan2018large,lee2019sumbt,TRADE} obtain the belief state directly from the dialogue history, combining NLU and DST. The word-level policy models \cite{MDRG,chen2019semantically,zhao2019rethinking} generate a natural language response according to the belief state and dialog history, combining POL and NLG.

\begin{figure}[t] 
\centering 
\includegraphics[width=0.7\linewidth]{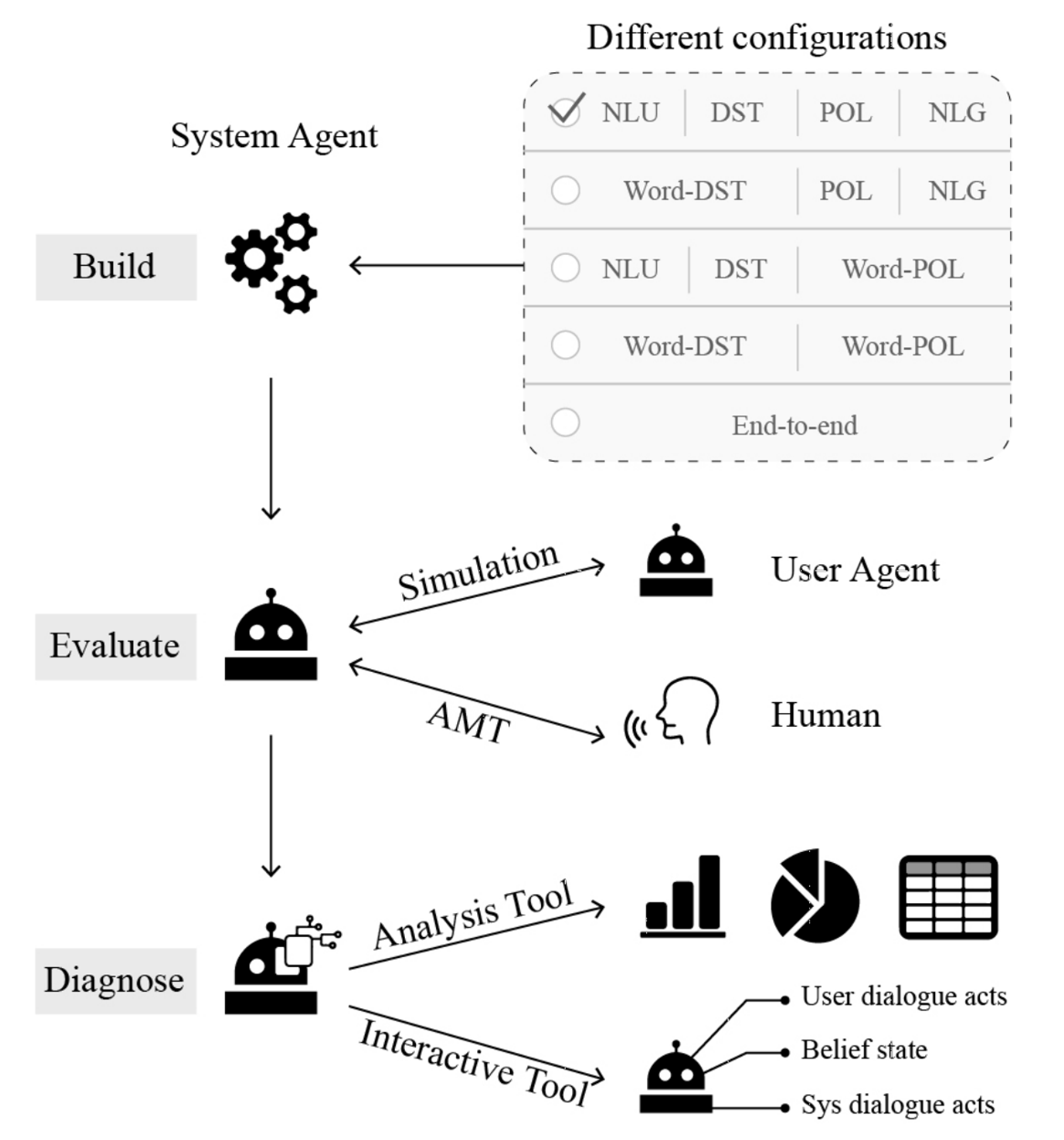}
\vspace{-1mm}
\caption{Framework of ConvLab-2, where the top block shows different system configurations for building a task bot \cite{zhu-etal-2020-convlab}.} 
\label{fig:convlab2} 
\vspace{0mm}
\end{figure}

Many of these models have been integrated into open-source platforms for building task bots, such as RASA \cite{rasa} and ConvLab \cite{lee2019convlab,zhu-etal-2020-convlab}, which allow developers to easily piece together task bots. 
As shown in Figure~\ref{fig:convlab2}, Convlab-2 provides a set of state-of-the-art modeling methods for different dialog modules. 
Researchers can quickly assemble, evaluate and debug a task bot by selecting a system configuration based on the complexity of the task, the availability of task-specific training data and so on. The task bot can be a classical pipeline system consisting of separately developed modules (as shown in the top row of the system configuration table in Figure~\ref{fig:convlab2}), a fully end-to-end system (the bottom row), or a joint-model system (the middle rows). 

These ML-based methods have succeeded in building state-of-the-art task bots 
for only a few domains that are well-studied in the research community.
However, they have not been widely adopted in developing commercial task bots because they require large amounts of fine-grained, task-specific labels for training \cite{takanobu2020your}, which are rarely available for many real-world tasks.
This is in contrast to the language model based approach, described in Section~\ref{sec:neural-language-model-bot}, where a large-scale neural language model can be pre-trained on open-domain datasets without task-specific labels.

Another reason that ML methods are not widely used is that it is difficult to maintain a commercial task bot by only using dialogs as training examples. Very often, it is far more effective to allow dialog authors to directly edit the dialog flow to accommodate the change of business rules. For example, the dialog authors may decide they want to change the flow pattern based solely on aesthetics or measures of deal flow. This type of change is different as it is not about adding additional training examples but about updating the ones that already exist to reflect the new preferred dialog path. Thus, a hybrid approach that combines dialog composers and ML methods is highly desirable. 

In the next section, we present a newly-developed hybrid approach that combines the best of the neural language model based approach and the classical modular approach for building robust task bots at scale.



\section{Grounded Text Generation}
\label{sec:gtg}







This section describes a hybrid approach, being developed contemporaneously by multiple research teams \cite{peng2020soloist,hosseini2020simple,Ham2020e2e}. 
It combines the strengths of the classical modular approach (Section~\ref{sec:modular-bot}) and the fully data-driven approach (Section~\ref{sec:neural-language-model-bot}). 
Our description follows closely \cite{peng2020soloist}, and we discuss, wherever appropriate, the differences in model design and implementation between \cite{peng2020soloist} and \cite{hosseini2020simple,Ham2020e2e}.

The hybrid approach aims to develop \emph{at scale} robust task bots whose behaviors are \emph{interpretable} and \emph{controllable}. 
To ensure interpretability and controllability, we follow the classical modular approach to architecting a task bot as a pipeline system where the output and input of each module in the pipeline is represented using symbols (such as slot-value pairs and templates) that are human comprehensible. 
To ensure scalability, the pipeline system is implemented using a stateful language model, called a Grounded Text Generation (GTG) model henceforth, which is equipped with an internal cognitive model of keeping track of dialog states, and can generate responses grounded in dialog states and task-specific knowledge. 
GTG is a hybrid model that uses Transformers \cite{transformer} as its backbone, combined with symbol-manipulation modules for knowledge base inference and prior knowledge encoding.
GTG can be pre-trained on open-domain datasets, and adapted to building bots for completing specific tasks with limited numbers of task labels. 

\subsection{GTG for task-oriented dialog}
\label{subsec:gtg4dialog}

\begin{figure}[t] 
\centering 
\includegraphics[width=0.99\linewidth]{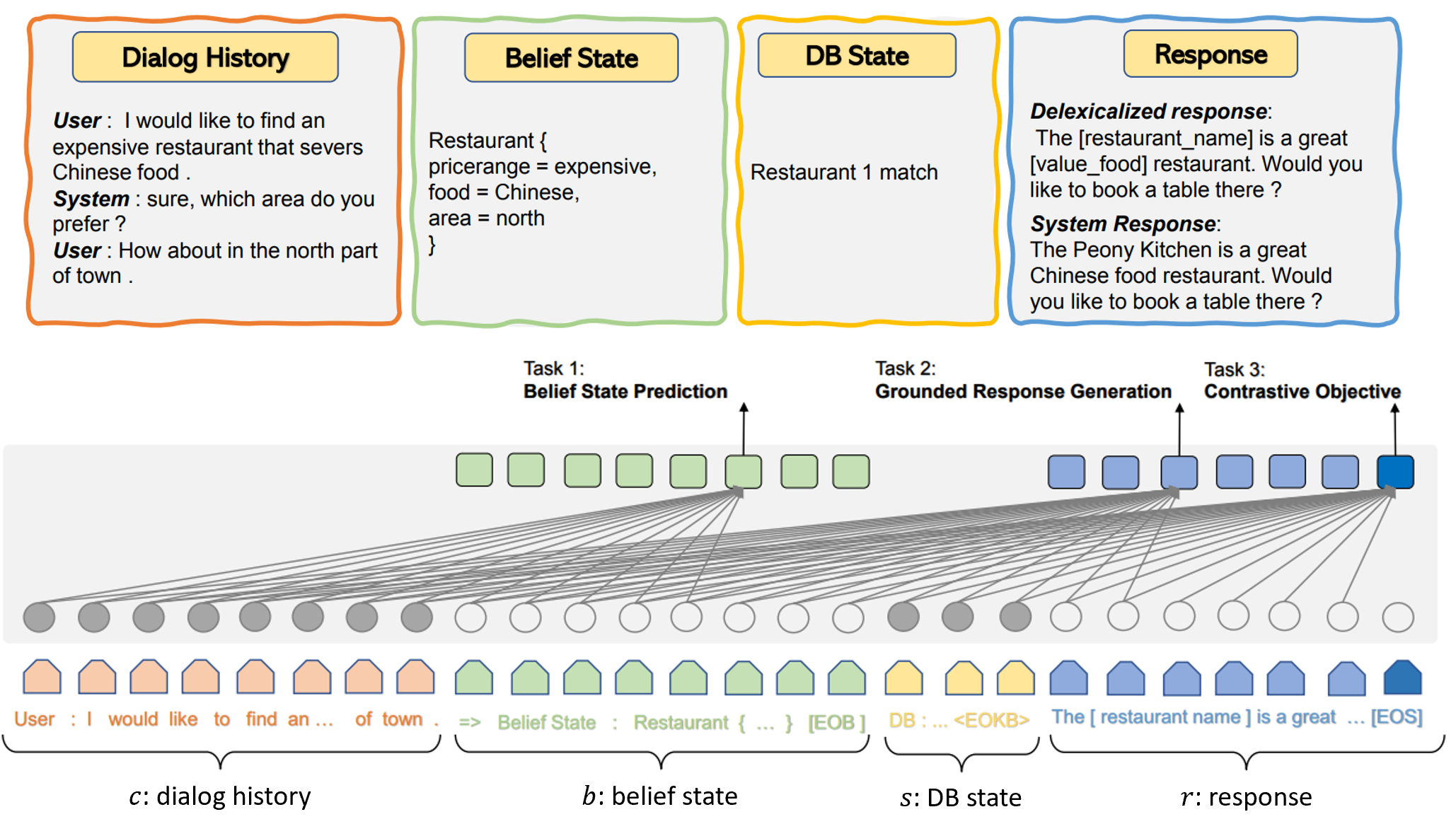}
\vspace{0mm}
\caption{GTG is an auto-regressive language model trained using multi-task learning. (Top) An example of a dialog turn. (Bottom) Multi-task learning of GTG using a dialog turn represented as a single text sequence in input and output. Adapted from \cite{peng2020soloist}.} 
\label{fig:gtg-mtl} 
\vspace{0mm}
\end{figure}

Consider a multi-turn task-oriented dialog, as illustrated in Figure~\ref{fig:gtg-mtl}. At each turn, GTG generates a natural language response $r$ in the following steps. 

\paragraph{State tracking.} 
Given the dialog history $c$, GTG generates a dialog brief state $b$, 
\begin{equation}
b=\text{GTG}(c)        
\label{eq:dst}
\end{equation}
where $b$ is a list of tuples recording values for slots in a domain (e.g., \texttt{(restaurant, pricerange=expensive)}). 
In this step, GTG functions like a word-DST model which obtains the belief state directly from the dialogue history, combining NLU and DST.

\paragraph{Knowledge base lookup.}
The belief state is used to query a task-specific knowledge base to retrieve the entities that satisfy the conditions specified in the belief state (e.g., the restaurants that meet the requirements on price, food type and location) as 
\begin{equation}
s=\text{GTG}(b)        
\label{eq:lookup}
\end{equation}
where $s$ is the database state, which includes the number of returned entities which is used later for response selection, and the attributes of each returned entity which are used later to lexicalize the response. In \cite{peng2020soloist}, database lookup is implemented using a keyword-matching-based lookup API. But this can also be implemented using neuro-symbolic retrieval models (e.g., \cite{iyyer2017search}), as will be described in Section~\ref{subsec:sym-modules}.

\paragraph{Grounded response generation.}
GTG generates a delexicalized response $r$ grounded in dialog belief state $b$, task-specific knowledge $s$ and dialog history $c$,
\begin{equation}
r=\text{GTG}(c,b,s)        
\label{eq:grg}
\end{equation}
where $r$ is delexicalized such that it contains only generic slot tokens (such as \texttt{[restaurant-name]} and \texttt{[value-food]}) rather than specific slot values. 
The final system response in natural language can be obtained by lexicalizing $r$ using information from $b$ and $s$. 
Compared to lexicalized responses, letting GTG generate delexicalized responses makes GTG more generic and its training more sample-efficient.  

The above generation process suggests that GTG can be viewed as an auto-regressive model, which can be implemented e.g., using a multi-layer Transformer neural network, parameterized by $\theta$. 
Representing each dialog turn as $x = (c, b, s, r)$, the joint probability $p_\theta(x)$ can be factorized as:
\begin{equation}
p_\theta(x) = p_\theta(c) p_\theta(b|c) p_\theta(s|b) p_\theta(r|c,b,s)        
\label{eq:gtg}
\end{equation}
where
$p_\theta(c)$ is the probability of the dialog history which can be computed recursively using Equation~\ref{eq:gtg},
$p_\theta(b|c)$ is the probability of belief state prediction as Equation~\ref{eq:dst},
$p_\theta(s|b)=1$ since according to Equation~\ref{eq:lookup}, the database state $s$ is obtained using a deterministic database lookup process given dialog belief state $b$, and
$p_\theta(r|c,b,s)$ is the probability of grounded response generation as Equation~\ref{eq:grg}. 

The model parameter $\theta$ can be optimized on training data $\mathcal{D}=\{ x_n\}_{n=1}^{N}$ using multi-task learning \cite{peng2020soloist}, as illustrated in Figure~\ref{fig:gtg-mtl}, where a dialog turn is presented using a simple text format, with special delimiter tokens to indicate the types of different segments, (e.g., \texttt{[EOB]} indicates the end of the belief state segment). 
In addition to belief prediction and grounded response generation, whose losses are defined as
\begin{equation}
\mathcal{L}_{\text{B}} = -\log(p_\theta(b|c)) 
\label{eq:loss-dst}
\end{equation}
and
\begin{equation}
\mathcal{L}_{\text{R}} = - \log(p_\theta(r|c,b,s)),
\label{eq:loss-rgr}
\end{equation}
the task of contrastive learning is added to to promote the matched items (positive samples $x$), while driving down mismatches (negative samples $x^{\prime}$). Specifically, a set of negative samples are sampled from $x$ by replacing some items in $x$ with probability 50\% with different items randomly sampled from the dataset $\mathcal{D}$. 
Since the the special token $\texttt{[EOS]}$ attends all tokens in the sequence, the output feature on $\texttt{[EOS]}$ is the fused representation of all items. We apply a binary classifier on top of the feature to predict whether the items of the sequence are matched ($y=1$) or mismatched ($y=0$):

\begin{equation}
\mathcal{L}_{\text{C}} = - y \log(p_\theta(x)) - (1-y) \log(1-p_\theta(x^{\prime})). 
\label{eq:loss-cl}
\end{equation}

Thus, for a training dataset $\mathcal{D}=\{x_n\}_{n=1}^N$ consisting of $N$ dialog turns (regardless of whether they are from the same dialog sessions or not),  the full training objective for multi-task learning is
\begin{equation}
\mathcal{L}_{\theta}(\mathcal{D}) = \sum_{n=1}^{N}
(\mathcal{L}_{\text{B}}(x_n) +
\mathcal{L}_{\text{R}}(x_n) +
\mathcal{L}_{\text{C}}(x_n)).
\label{eq:loss-mtl}
\end{equation}

The models proposed in \cite{Ham2020e2e,hosseini2020simple} differ from the GTG model described above (which follows \cite{peng2020soloist}) in that the response generation in \cite{Ham2020e2e,hosseini2020simple} is performed in two steps, following the classical pipeline architecture where POL and NLG are performed by two separate modules.
Instead of generating $r$ directly from $(c, b, s)$ as in Equation~\ref{eq:grg}, system action $a$ is first generated based on $(c, b, s)$, and delexicalized response $r$ is generated based on $(c, b, s, a)$ as:

\begin{equation}
a=\text{GTG}(c,b,s)        
\label{eq:gra}
\end{equation}
and
\begin{equation}
r=\text{GTG}(c,b,s,a).        
\label{eq:nlg}
\end{equation}

The authors of \cite{peng2020soloist} argue that combining POL and NLG is critical to make the approach scalable. 
By separating POL and NLG, the system actions need to be labeled for training. 
However, system actions are task-specific labels, and are not easy to be collected in large amounts. 
The only labels needed for training GTG, on the other hand, are belief states, which are not only relatively task-independent but similar to named entity annotations that can be collected in large quantities much more easily, e.g., by transforming Wikipedia documents \cite{nothman2008transforming}.

Therefore, while the models of \cite{Ham2020e2e,hosseini2020simple} need to be trained on labeled, task-specific dialog data, one for each task, GTG can be pre-trained on open-domain datasets and then fine-tuned for specific tasks using much less task labels. For example, in \cite{peng2020soloist} the parameters of GTG model are initialized using GPT-2, and then pre-trained on large heterogeneous dialog corpora using multi-task learning according to the loss of Equation~\ref{eq:loss-mtl}.

\subsection{Symbol-Manipulation Modules in GTG}
\label{subsec:sym-modules}

The symbol-manipulation modules in GTG allow bot developers to incorporate task-specific code and business rules, and access task-specific knowledge bases.  
This section describes these modules, including a task-specific named entity recognizer, action masks that indicate actions which are permitted or not permitted at the current dialog state based on, e.g., business rules \cite{williams2017hybrid}, response templates that give detailed control over the way a task bot respond to users,
and a question answering (QA) module that can retrieve from a task-specific knowledge base the entities that meet users' requests. 

\paragraph{Task-specific named entity recognizer.}
Although the pre-trained GTG can identify many common types of named entities for dialog state tracking, there are always task-specific named entities that are unseen in data used for its pre-training. 
Thus, it is desirable to allow dialog developers to plug-in a task-specific named entity recognizer (e.g., which can be implemented based on table lookup, or a machine-learned model, or a mix of both) to identify entity mentions in dialog history $c$, which can be passed to GTG for state tracking. Equation ~\ref{eq:dst} can then be extended to $b=\text{GTG}(c, e)$, where $e$ is the list of tuples consisting of task-specific named entity mentions and their positions in $c$, and GTG can be extended by incorporating the copying mechanism \cite{gu2016incorporating} to help place $e$ into proper places in $b$. 

\paragraph{Action mask.}
An action mask can be used to explicitly encode commonsense rules or business rules, which are hard, or unnecessary, to learn from data.
For example, in a customer support bot \cite{williams2017hybrid}, if a target phone number has not yet been identified in the current dialog state, the API action to place a phone call is masked according to commonsense.
In a sales bot, a returned customer is always provided with a coupon during holiday seasons according to the business policy.  
The action mask proposed in \cite{williams2017hybrid} is only applicable to task bots where their action space is bounded. 
These bots can only respond by choosing one of a set of a finite number of actions, defined as e.g., action templates. 
The GTG described in Section \ref{subsec:gtg4dialog}, however, assumes an unbounded action space. This could be resolved by fine-tuning GTG to a task whose action space is bounded, as we will discuss in detail in Section~\ref{subsec:gtg-finetuning}.

\paragraph{Response template.}
GTG uses a language model to generate arbitrary system responses as in Equation~\ref{eq:grg} (or Equation~\ref{eq:nlg}), but does not have very detailed control over exactly what a response is composed of. Traditional task bots use template-based NLG to give good control over the system responses, but it is time-consuming and error-prone to manually define templates to cover all possible system responses.

\begin{figure}[t] 
\centering 
\includegraphics[width=0.99\linewidth]{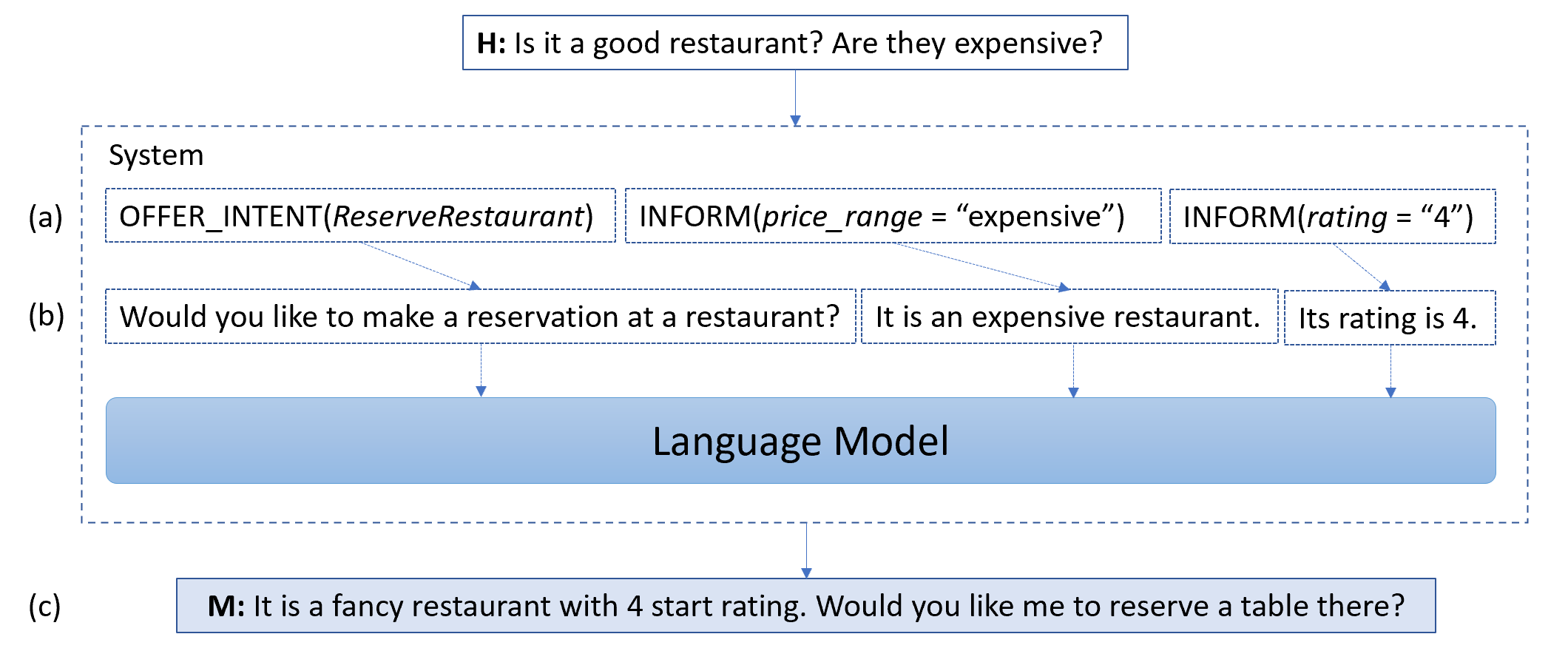}
\vspace{-1mm}
\caption{An example of template rewriting for response generation, adapted from \cite{kale2020few}: (a) The policy outputs a set of actions in response to the user input. (b) Simple templates convert each action into a natural language utterance. (c) Template-generated utterances are concatenated and fed to a language model to generate a system response.} 
\label{fig:template-rewriting} 
\vspace{0mm}
\end{figure}

A template rewriting method that combines templates and a language model is proposed in \cite{kale2020few}. As illustrated in Figure~\ref{fig:template-rewriting}, the method employs a set of simple templates to convert actions into utterances, which are concatenated to give a semantically correct, but possibly incoherent and ungrammatical utterance. Then a language model rewrites it into a coherent and fluent natural language response.  
The authors show that the templates are used as simple representations of actions to assist response generation, and do not need to cover all edge cases typically required in traditional template-based methods e.g., handling of plurals, subject-verb agreement, morphological inflection etc. For most tasks, 15 to 30 templates are enough. 

\begin{figure}[t] 
\centering 
\includegraphics[width=0.7\linewidth]{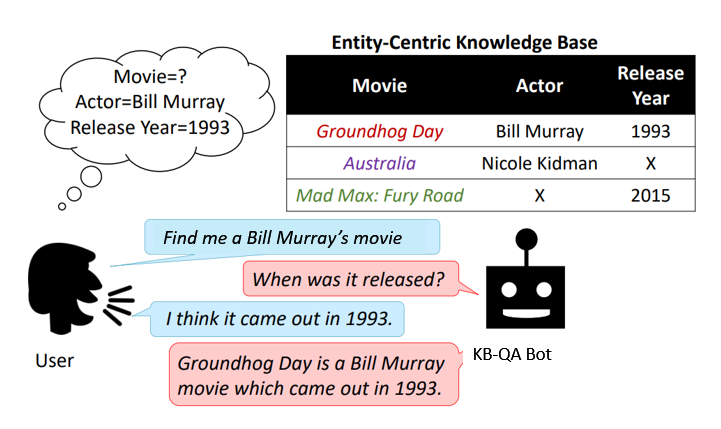}
\vspace{-1mm}
\caption{An interaction between a user and a knowledge base QA bot for the movie-on-demand task~\cite{dhingra17towards}.} 
\label{fig:movie-on-demand-example} 
\vspace{0mm}
\end{figure}

\paragraph{QA over knowledge bases.}
Many task bots are developed as a natural language interface to access their task-specific entity-centric knowledge bases (KBs), where a list of entities are stored, each associated with a set of properties. 
A KB is typically stored as a relational database or a knowledge graph.  
Relational datasets predominate because they use less storage space and are much faster than knowledge graphs when operating on a large number of records. 
An example of movie-on-demand bot and its KB is illustrated in Figure~\ref{fig:movie-on-demand-example}.

Accessing KB via a natural language or conversational interface is an active research field. While table lookup based on keyword matching is still widely used in many task bots, there is a growing interest in developing hybrid neuro-symbolic models for QA over knowledge bases \cite{iyyer2017search,yih2015semantic,guo2018dialog}.
These models are typically known as \emph{semantic parsers}. 
They map a dialog history (or a dialog state) to a meaning representation in formal logic, which can be executed on the KB to produce the answer.

Consider the example in Figure~\ref{fig:movie-on-demand-example}. 
Given the dialog history up to user's input ``I think it came out in 1993'', the meaning representation generated by a semantic parser is ``Select \textbf{Movie} Where \{\textbf{Actor} = Bill Murray\} AND \{\textbf{Release Year} = 1993\}'', and its returned result is \{Goundhog Dog\}.  
The meaning representation in this example is a SQL-like query, which consists of a ``select'' statement that is associated with the name of the answer column, and zero or more conditions, each containing a condition column and an operator (=, <, >) and arguments, which enforce additional constraints on which cells in the answer column can be chosen.

In what follows, we use the Dynamic Neural Semantic Parser (DynSP) \cite{iyyer2017search} as an example to illustrate how the semantic parsing problem is formulated and dealt with using a hybrid neuro-symbolic approach.
Given a dialog history and a table that stores the entities and their associated properties, DynSP formulates semantic parsing as a state-action search problem, 
where a state $\mathcal(S)$ is defined as an action sequence representing a complete or partial parse, and an action $\mathcal{A}$ is an operation to extend a parse. 
Parsing is cast as a process of searching an end state with the highest score.

\begin{figure}[t] 
\centering 
\includegraphics[width=0.6\linewidth]{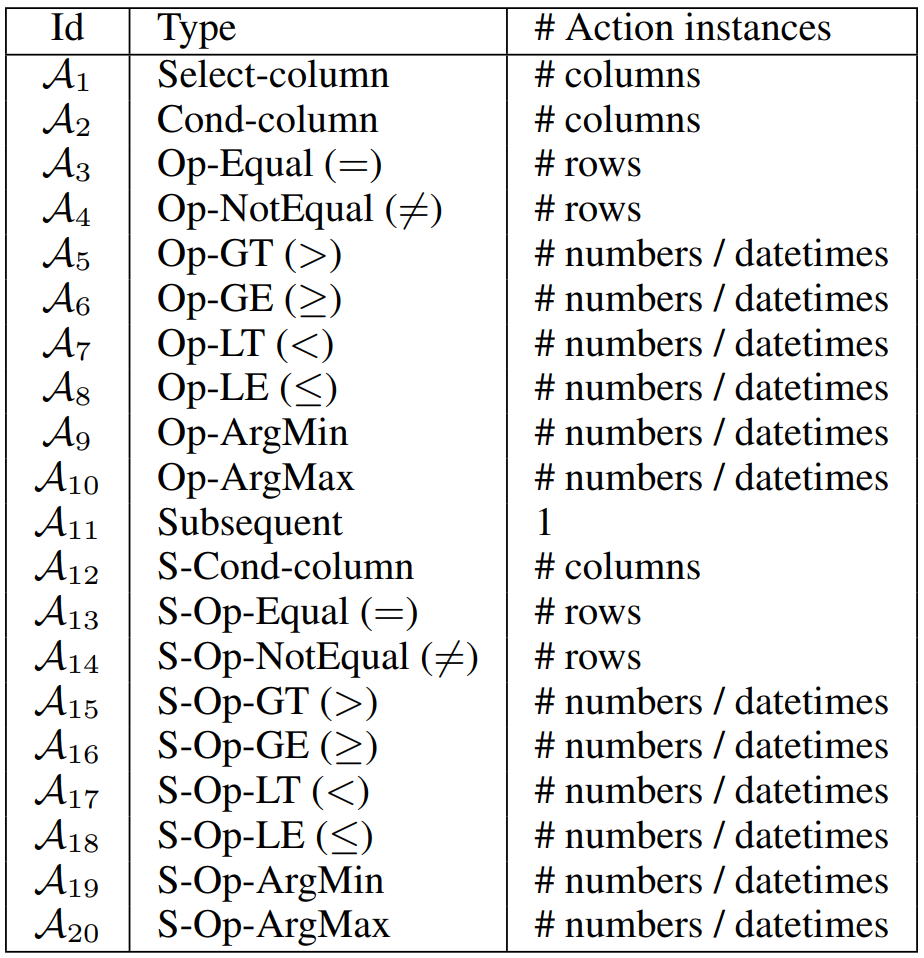}
\vspace{-1mm}
\caption{Types of actions and the number of action instances in each type, defined in \cite{iyyer2017search}.} 
\label{fig:dynsp-actions} 
\vspace{0mm}
\end{figure}

DynSP is inspired by STAGG, a search-based semantic parser \cite{yih2015semantic} and the dynamic neural module network (DNMN) \cite{andreas2016learning}.  
Like STAGG, DynSP pipelines a sequence of modules as search progresses; but these modules are implemented using neural networks, which enables end-to-end training as in DNMN. 
Note that in DynSP the network structure is not predetermined, but are constructed dynamically as the parsing procedure explores the state space. 
Figure~\ref{fig:dynsp-actions} shows the types of actions and the number of action instances in each type, defined in \cite{iyyer2017search}. 
Consider the example in Figure~\ref{fig:movie-on-demand-example}, one action sequence that represents the parser is 
\{($\mathcal{A}_1$) select-column \textbf{Movie}, ($\mathcal{A}_2$) cond-column \textbf{Actor}, ($\mathcal{A}_3$) op-equal ``Bill Murray'', ($\mathcal{A}_2$) cond-column \textbf{Release Year}, ($\mathcal{A}_5$) op-equal ``1993''\}.

\begin{figure}[t] 
\centering 
\includegraphics[width=0.5\linewidth]{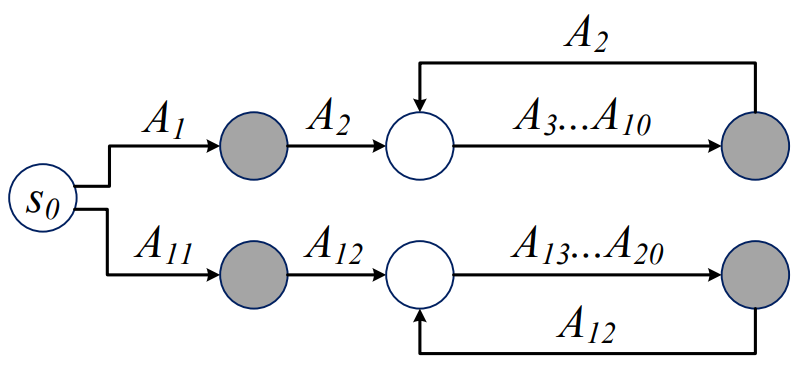}
\vspace{-1mm}
\caption{Possible action transitions based on their types (see Figure~\ref{fig:dynsp-actions}), where shaded circles are end states \cite{iyyer2017search}.} 
\label{fig:dynsp-fsm} 
\vspace{0mm}
\end{figure}

Since many states represent semantically equivalent parses, to prune the search space, the actions that can be taken for each state are pre-defined, as shown in Figure~\ref{fig:dynsp-fsm}. 
Then, beam search is used to find an end state with the highest score in the space using the state value function as: 

\begin{equation}
V_\theta (s_t) = V_\theta (s_{t-1}) + \pi_\theta (s_{t-1}, a_t), V_\theta (s_0) = 0 
\label{eq:dynsp}
\end{equation}
where $s_t$ is a state consisting of an action sequence $a_1,...a_t$, and $\pi(s,a)$ is the policy function that scores action $a$ given state $s$. 

Equation~\ref{eq:dynsp} shows that the state value function can be decomposed as a sequence of policy functions, each implemented using a neural network. 
Therefore, the state value function is a state-specific neural network parameterized by $\theta$ which can be trained end-to-end. \cite{iyyer2017search} propose to learn $\theta$ using weakly supervised learning on query-answer pairs without ground-truth parses. 
This is challenging because the supervision signals (rewards) are delayed and sparse. E.g., whether the generated parse is correct or not is only known after a \emph{complete} parse is generated and the KB lookup answer is returned. 
The authors use an approximate reward, which is dense, for training. 
A \emph{partial} parse is converted to a query to search the KB, and the overlap of its answers with the gold answers is used as the training supervision: A higher overlap indicates a better partial parse.


\subsection{Adapting GTG to Specific Tasks}
\label{subsec:gtg-finetuning}

When deploying the pre-trained GTG model to a specific task, we need to adapt GTG so that the entity mentions detected in user utterances can be grounded in task-specific knowledge base instances, and the generated responses are not only optimized for task completion but consistent with task-specific business rules. 
The adaptation can be achieved using methods of supervised learning, reinforcement learning, or the combination of both.

\subsubsection{Supervised Learning}
\label{subsubsec:supervised-learning}

This section describes three methods of adapting GTG to a new task using annotated task-specific dialog logs, a dialog flow, and machine teaching, respectively.

These are supervised learning methods in that they all optimize GTG for mimicking human-agent policies (behaviors) of completing a task. The human policies are presented in the forms of annotated human dialog logs or manually-crafted dialog flows, or are explicitly provided by human teachers (dialog authors) through a machine teaching tool.

\begin{figure}[t] 
\centering 
\includegraphics[width=0.9\linewidth]{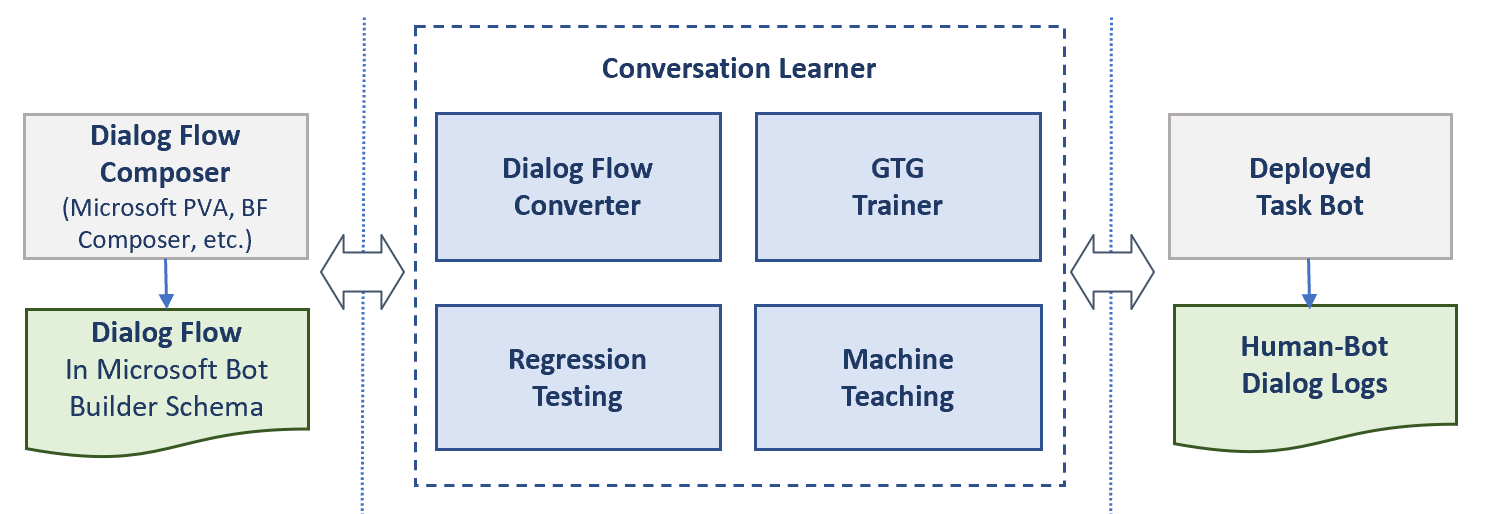}
\vspace{-1mm}
\caption{Architecture of conversation learner, adapted from \cite{shukla2020conversation}. (Left) Dialog flows in Microsoft `.dialog' schema 
developed manually by dialog authors using a dialog flow composer tool. (Middle) Four components of conversation learner. (Right) Human-bot dialog logs collected via the interactions between users and a deployed task bot.} 
\label{fig:cl} 
\vspace{0mm}
\end{figure}

We will use Conversation Learner (CL), a machine teaching tool for building task bots \cite{shukla2020conversation}, as an example to illustrate how these methods are implemented. The overall architecture of CL is shown in Figure~\ref{fig:cl}. 
It consists of four components: 
(1) a dialog flow converter that converts a dialog flow, which is manually created by dialog authors using a dialog flow composer (e.g., Microsoft's PVA), to a set of annotated dialogs for GTG adaptation,
(2) a GTG trainer that adapts a pre-trained GTG model to a task using annotated task-specific dialog logs, 
(3) a machine teaching module that allows dialog authors to interactively \emph{teach} the bot how to complete the task by correcting the mistakes made by the bot, and
(4) a regression testing module that allows side-by-side comparison of the dialogs (stored in a regression test set) generated by different versions of GTG being adapted, ensuring that the bot learns new skills without forgetting previously learned skills that are considered worth memorizing.

\footnotetext{{https://github.com/microsoft/botbuilder-dotnet/blob/master/schemas/sdk.schema}}

\paragraph{Adaptation using dialog logs.}
If annotated dialog logs are available for a new task, we can use the conventional fine-tuning procedure to adapt GTG. Assuming that the log data is in the form of a set of $x = (c, b, s, r)$ as in Equation~\ref{eq:gtg}, where the representation of each dialog turn consists of dialog context $c$, delexicalized response $r$, labeled by dialog belief state $s$ and database state $s$, as illustrated by the example in Figure~\ref{fig:gtg-mtl}, we can use the same multi-task objective of Equation~\ref{eq:loss-mtl} to update $\theta$ to adapt the model to complete the new task. 

For a task with a $K$-bounded action space (i.e. the bot is allowed to respond using one of $K$ pre-defined action templates), we can replace $r$ in $x$ with a template-ID. That is, we convert the grounded response generation task of Equation~\ref{eq:grg} to $K$-way classification. Action masks can be applied in the bots with bounded action space to indicate at each dialog turn which actions are (not) permitted based on task-specific business rules. Specifically, an action mask is applied at the softmax layer of GTG (for response generation) to mask all the actions that are not permitted given the current dialog belief state, and the result is re-normalized to select the highest-probability action as the response for the current turn.

Using CL, dialog authors can simply upload the annotated dialog logs and fine-tune GTG using the GTG trainer 
\footnote{If masks are employed, dialog authors must also specify per-action masks.}. 

\begin{figure}[t] 
\centering 
\includegraphics[width=0.8\linewidth]{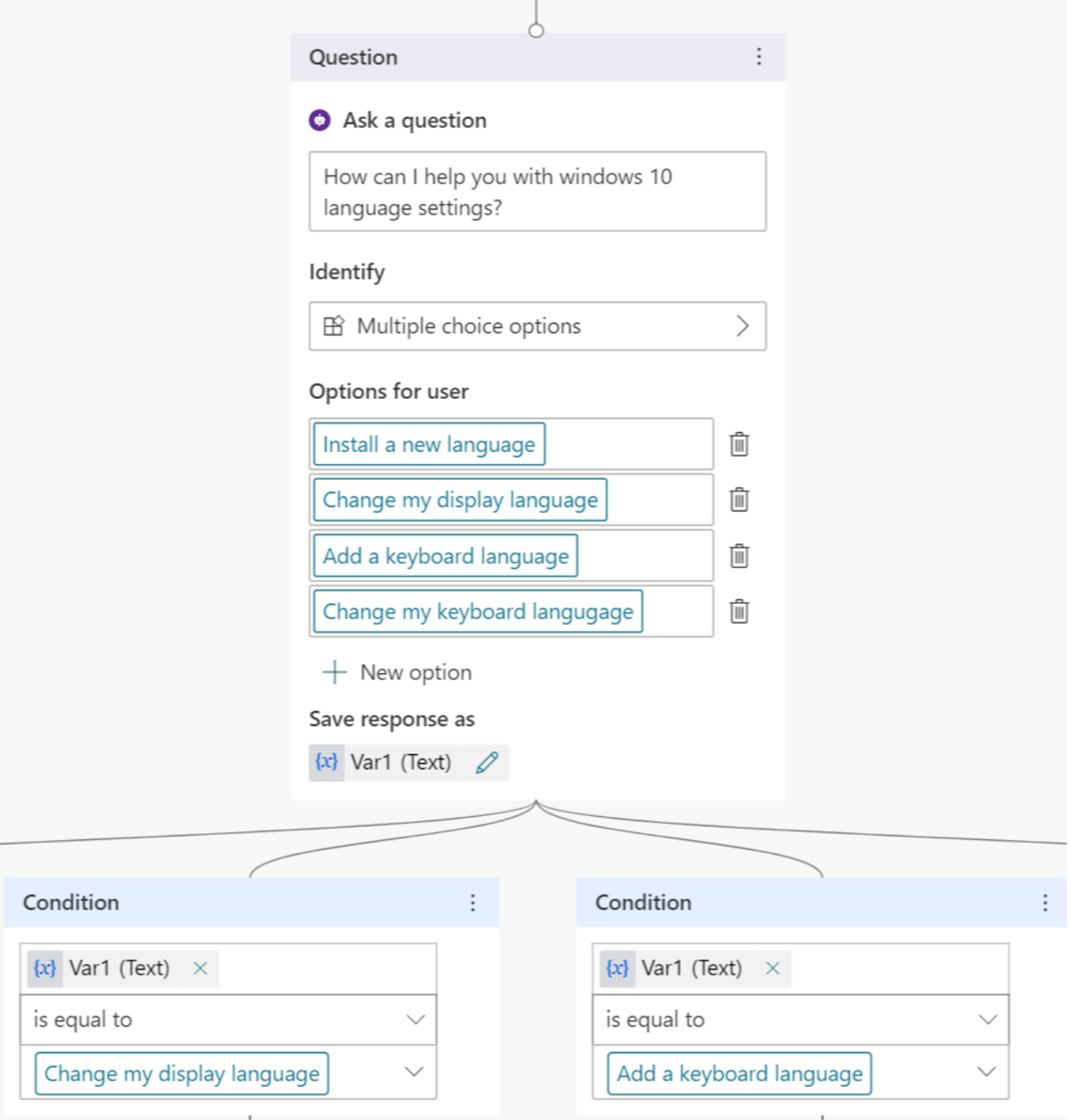}
\vspace{-1mm}
\caption{An example of a dialog flow composed in the Microsoft PVA system \cite{shukla2020conversation}.} 
\label{fig:pva-dialog-flow} 
\vspace{0mm}
\end{figure}

\paragraph{Adaptation using a dialog flow.}
In a typical industrial implementation of task bots, the dialog strategy is expressed as a dialog flow, which is a finite state machine, composed manually using dialog composers. Figure~\ref{fig:pva-dialog-flow} shows a dialog flow composed using Microsoft's PVA dialog composer.

\begin{figure}[t] 
\centering 
\includegraphics[width=0.98\linewidth]{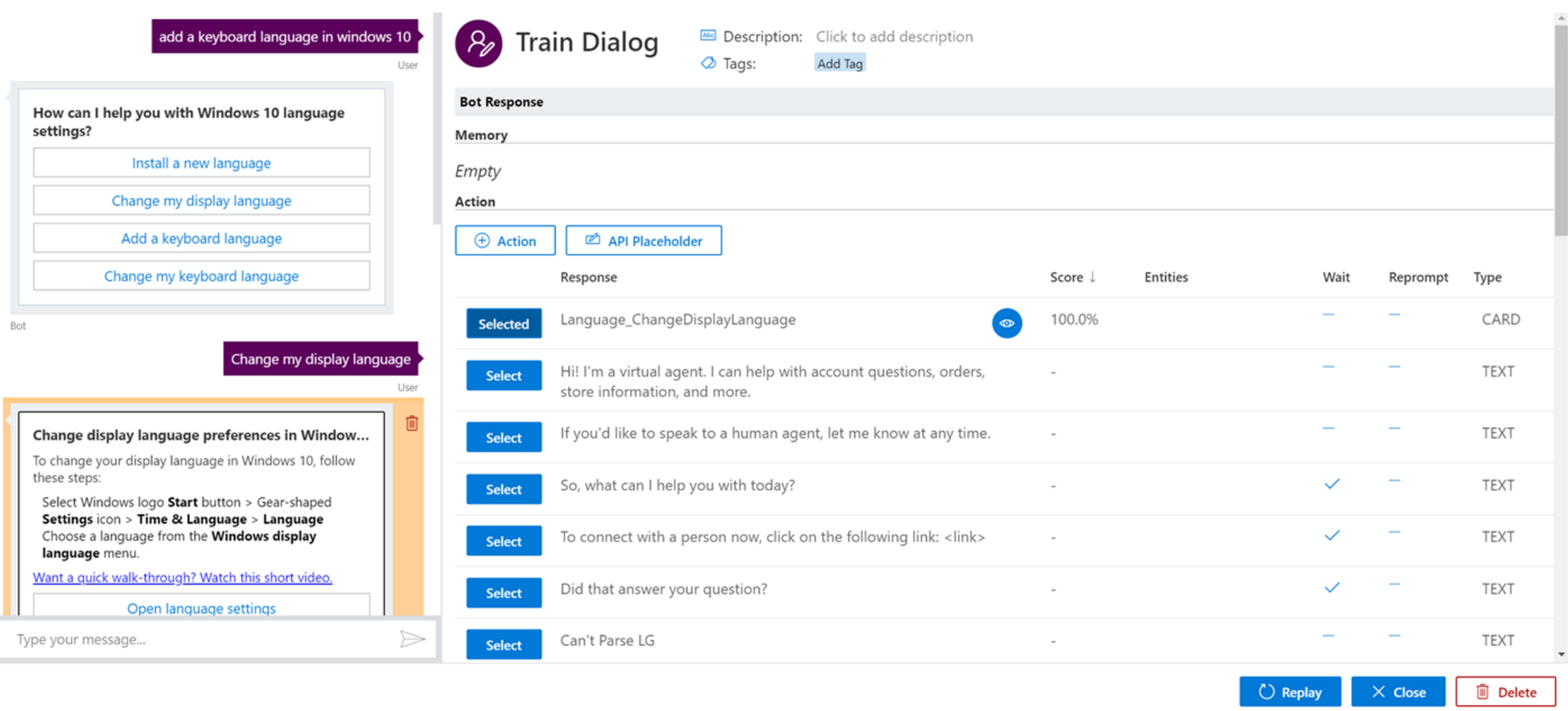}
\vspace{-1mm}
\caption{An example of a training dialog generated from the dialog flow of Figure~\ref{fig:pva-dialog-flow} using the dialog flow converter of CL by traversing different paths of the dialog history (left) and actions (right). Adapted from \cite{shukla2020conversation}.} 
\label{fig:cl-dialog-flow} 
\vspace{0mm}
\end{figure}

GTG can be adapted to a task using its dialog flow, if it is available, in two steps using CL. 
In the first step, the dialog flow converter automatically generates a set of training dialogs that represent the dialog flow. This is done by performing an exhaustive set of walks over the dialog flow and generating training dialog instances for each walk. Rules that determine transitions in the dialog flow are represented as action masks. Figure~\ref{fig:cl-dialog-flow} shows an example of a generated training dialog from the dialog flow shown in Figure~\ref{fig:pva-dialog-flow}. Since in the dialog flow, bot's action template $r$ and dialog state $(b, s)$ are explicitly defined given dialog history $c$, the generated dialogs are labeled training samples of the form $x = (c, b, s, r)$.
In the second step, GTG is fine-tuned to the task using the generated dialogs.

\paragraph{Adaptation using machine teaching.}
Machine teaching is an active learning paradigm that leverages the expertise of domain experts as \emph{teachers}. The machine teaching module in CL allows teachers (dialog authors) to select and visualize dialogs, find potential problems, and provide corrections or additional training samples to improve the bot's performance.
A pre-trained GTG is adapted to a specific task using CL in the following steps:
\begin{enumerate}
  \item Dialog authors deploy a pre-trained GTG for a specific task.
  \item Users (or human subjects recruited for system fine-tuning) interact with the bot and generate human-bot dialog logs.
  \item Dialog authors select representative failed dialogs from the logs and correct their belief states and/or responses so that the bot can complete these dialogs successfully. These corrections are saved as training dialogs used to incrementally adapt GTG to the task.
\end{enumerate}

\begin{figure}[t] 
\centering 
\includegraphics[width=0.99\linewidth]{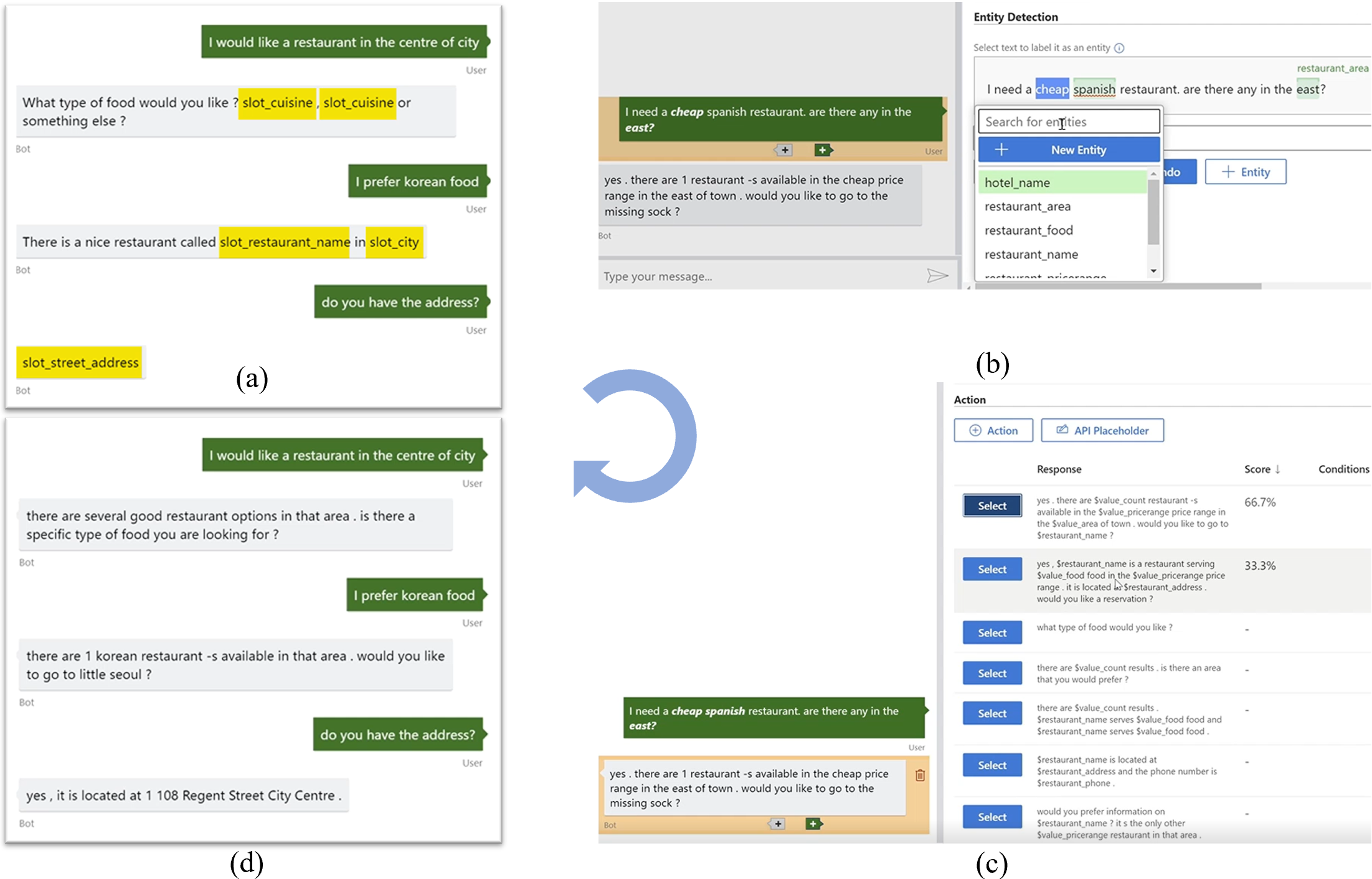}
\vspace{-1mm}
\caption{Illustration of the machine teaching process using a restaurant booking task as an example. (a) shows a conversation between a user and a pre-trained GTG. Dialog authors then incrementally review the logged dialogs, fine-tune the GTG by labeling slots (b) and correcting responses (c). (d) is the same conversation shown in (a) with the fine-tuned GTG.} 
\label{fig:cl-example} 
\vspace{0mm}
\end{figure}

Figure~\ref{fig:cl-example} illustrates the machine teaching process using a restaurant booking task as an example.
Figure~\ref{fig:cl-example}(a) shows a conversation between a user and a pre-trained GTG. We see that  the pre-trained GTG model already has knowledge about entities and slots for restaurant booking and can generate proper delexicalized responses for task completion. However, it has not been grounded in any specific restaurant booking database instance and it only generates responses with unfilled slot values based on dialog context. 
Dialog authors then incrementally review the logged dialogs, labeling slots (b) and correcting responses (c). These corrections are saved as training dialogs that are used to incrementally adapt GTG to the task.
The dialog shown in (d) is the same sample conversation in (a) with the fine-tuned GTG. We see that using a few training dialogs for model fine-tuning, GTG can produce grounded responses that achieve the user goal. 

\paragraph{Human-in-the-loop adaptation.}
Some customers always want a human in the loop for providing services. The task bot is employed not to replace the human agent but to reduce human agent training time and to decrease human agent response latency. 
In this scenario, the bot suggests possible top responses for the human agent. 
On each turn, the human agent can pick from one of the suggested responses or add a new one. These responses can be further used to train the bot.

\subsubsection{Reinforcement Learning}
\label{subsec:reinforcement-learning}

A task-oriented dialog can be formulated as a decision making process under the reinforcement learning (RL) framework \cite{young2013pomdp,gao2019neural}. The task bot navigates in a Markov Decision Process (MDP), interacting with its environment (e.g., users and task-specific knowledge databases) over a sequence of discrete steps. At each step, the bot observes the current state, chooses an action according to a policy, and then receives a reward and observes a new state, continuing the cycle until the episode (dialog) terminates. The goal of reinforcement learning is to find the optimal policy to maximize expected rewards.

The RL setting differs from supervised learning in two aspects. First, while in supervised learning the bot learns to mimic human responses by following human (teacher) examples explicitly presented in the labeled training data, in RL the bot learns how to respond by exploring the state space and collecting reward signals by interacting with users (or user simulators). Second, unlike supervised learning where the labels are available for each sample, the reward signals in RL are often sparse and delayed, e.g., whether a task is completed or failed is known only after the dialog terminates.   

The multi-task learning of GTG, as described in \ref{subsubsec:supervised-learning}, is supervised learning. Now, we describe RL methods that might be applied to adapt GTG to specific tasks.

The objective of RL is to maximize the expected reward over the dialogs generated by a user and the target task bot. Formally, the objective is 
\begin{equation}
J(\theta) = \mathbb{E}
[R(x_1,x_2,...,x_T)] 
\label{eq:rl-objective}
\end{equation}
%
where $R(.)$ is the reward function. The reward can be defined according to task success rate. For example, a successful dialog corresponds to a reward of 30, a failure to a reward of -10, and we assess a per turn penalty of -1 to encourage pithy exchanges. 

The objective can be optimized using gradient descent by factoring the log probability of the conversation and the accumulated reward, which is independent of the model parameters:

\begin{equation}
\begin{split}
J(\theta) & = \nabla \log p_\theta(x_1,x_2,...,x_T) R(x_1,x_2,...,x_T) \\
          & \simeq \nabla \log \prod_t 
          p_\theta(x_t) R(x_1,x_2,...,x_T)
\end{split}
\label{eq:rl-sgd}
\end{equation}
%


where $p_\theta(x)$ is parameterized the same way of the Transformer-based auto-regressive model of Equation \ref{eq:gtg}, except that $p_\theta(x)$ in Equation \ref{eq:rl-sgd} is viewed as a dialog policy that indicates how likely response $r$ is selected at each dialog turn, and that $\theta$ is optimized using RL. In practice, learning a good policy from scratch is challenging. Thus, the model trained using supervised learning as Equation \ref{eq:loss-mtl} can be used as the initial policy for RL.

RL allows a task bot to learn how to respond in an environment which is different from the one where training data is collected. This is desirable since after we deploy bots to serve users, there is often a need over time to adapt to a changing environment (e.g., due to the need of adding user intents and task slots) \cite{lipton2018bbq,gavsic2014incremental}. 
In addition to using supervise learning and machine teaching for adaptation as described in \ref{subsubsec:supervised-learning}, RL provides an alternative solution for a bot to adapt without a teacher but from data collected by directly interacting with users in an initially unknown environment. 

In general, the bot has to try new actions (responses) in novel states to discover potentially better policies. Hence, it has to strike a good trade-off between exploitation (choosing good actions to maximize reward using the policy learned so far) and exploration (choosing new actions to discover better alternatives), leading to the need for efficient exploration \cite{sutton2018reinforcement}. 
Since $p_\theta(r|c,b,s)$ in Equation \ref{eq:gtg} is the probability distribution over all actions, we can view $p_\theta(x)$ as a stochastic policy that allows the bot to explore the state space without always taking the same action, thus handling the exploration-exploitation trade-off without hard coding it.

\begin{figure}
\centering
\includegraphics[width=\textwidth]{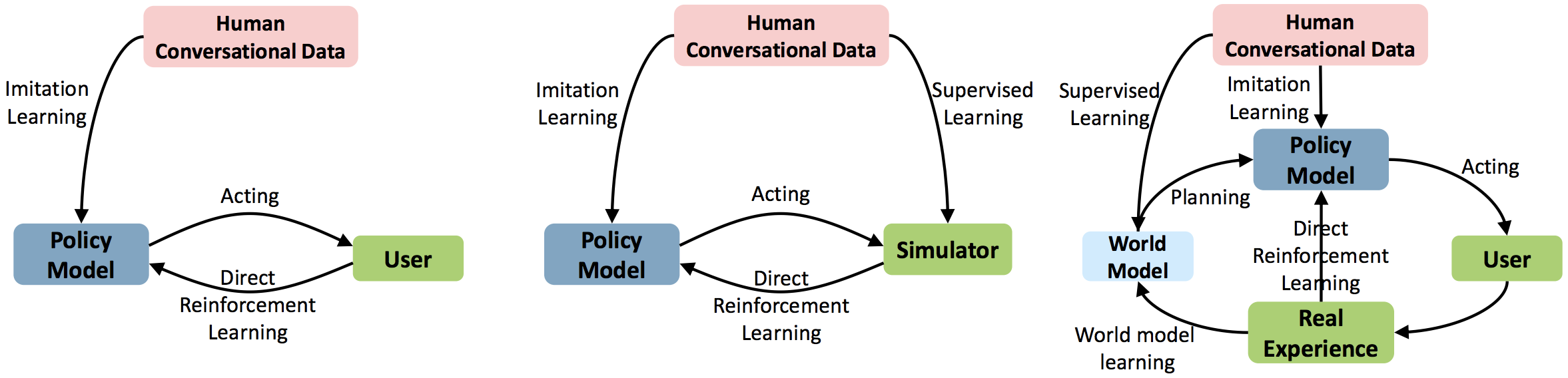}
\caption{Three strategies for optimizing dialog policies based on reinforcement learning (RL) \cite{peng2018deep} by interacting with human users (Left), user simulators (Middle), and both human users and user simulators (Right), respectively.} 
\label{fig:deep-dyna}
\end{figure}

Using RL for adaptation is compelling in theory since it needs neither pre-collected labeled data nor human teachers. But optimizing a task bot against human users is costly as it requires many interactions between the bot and humans (Figure \ref{fig:deep-dyna} (Left)). User simulators provide an inexpensive alternative to RL-based policy optimization (Figure \ref{fig:deep-dyna} (Middle)). The user simulators, in theory, do not incur any real-world cost and can provide unlimited simulated experience for RL. But user simulators usually lack the conversational complexity of human interlocutors, and the trained task bot is inevitably affected by biases in the design of the simulator. \cite{dhingra17towards} demonstrates a significant discrepancy in a simulator-trained task bot when evaluated with simulators and with real users.

Inspired by the Dyna-Q framework \cite{sutton1990integrated}, Peng et al. \cite{peng2018deep} propose Deep Dyna-Q (DDQ) to deal with large-scale RL problems with deep learning models. As shown in Figure \ref{fig:deep-dyna} (Right), DDQ allows a bot to optimize the policy by interacting with both human users and user simulators. Training of DDQ consists of three parts:
\begin{itemize}
\item{\emph{direct reinforcement
learning}: the dialogue system interacts with a human user, collects real dialogs and improves the policy by either imitation learning or reinforcement learning;}
\item{\emph{world model learning}: the world model (user simulator) is refined using real dialogs collected by direct reinforcement learning;}
\item{\emph{planning}: the dialog policy is improved against user simulators by reinforcement learning.}
\end{itemize}
Human-in-the-loop experiments show that DDQ is able to efficiently improve the dialog policy by interacting with real users, which is important for deploying dialog systems in practice \cite{peng2018deep}.

Several variants of DDQ are proposed to improve the learning efficiency. 
\cite{su2018discriminative} propose the Discriminative Deep Dyna-Q (D3Q) that is inspired by generative adversarial networks to better balance samples from human users and user simulators. Specifically, it incorporates a discriminator which is trained to differentiate experiences of user simulators from those of human users. During the planning step, a simulated experience is used for policy training only when it appears to be a human user experience according to the discriminator.
Similarly, \cite{wu2019switch} extend DDQ by integrating a \emph{switcher} that automatically determines whether to use a real or simulated experience for policy training.
\cite{zhang2019budgeted} propose a Budget-Conscious Scheduling-based Deep Dyna-Q (BCS-DDQ) to best utilize a fixed, small number of human interactions (budget) for task-oriented dialog policy learning. They extend DDQ by incorporating budget conscious scheduling to control the budget and improve DDQ’s sample efficiency by leveraging active learning and human teaching.

Although RL has not be widely used in building task bots for real-world applications, it is an active research topic and can potentially be combined with supervised learning for GTG adaptation.  

\subsection{Remarks on Continual Learning}
\label{subsec:continual-learning}

The capability of continual learning is important for a task bot to continuously improve its performance and adapt to a changing environment after its deployment.  
The environment could change due to various reasons. There may be a need over time to extend the domain of the target task by adding intents and slots \cite{gavsic2014incremental,lipton2018bbq}, to serve a new group of users whose behaviors are different from the users based on which the bot is originally designed and trained, or to update knowledge bases which might include incomplete or erroneous information.
We are implementing three primary continual learning skills in GTG: continual machine teaching, continual learning from user interactions, and updating knowledge for task completion.

\paragraph{Continual machine teaching.}
This is achieved by using conversation learning (CL), as described in Section~\ref{subsubsec:supervised-learning}. After the deployment of a bot, CL stores user-system dialog sessions in the logs and selects a set of representative (failed) dialog sessions for dialog authors to teach the bot to correct its behaviors so that the bot can successfully complete the same or similar tasks next time. 
It is also crucial to avoid catastrophic forgetting \cite{french1999catastrophic,kirkpatrick2017overcoming}
in which a bot improved using new dialog logs forgets how to deal with previously successfully handled tasks. This can be prevented by storing a collection of dialog sessions generated by previous versions of GTG that are considered representative (the regression dataset). By always re-training GTG using the regression dataset we can ensure that the bot learns new skills without forgetting previously learned skills.  
But how to select dialog sessions for the regression set and how to use the regression set for machine teaching are open research problems.

The change of business rules over time can also be handled by using CL. This type of change is typically made by dialog authors directly changing the dialog flow using dialog composers. Then, GTG can be adapted to the changed dialog flow using the dialog flow converter and the GTG trainer in CL as described in Section~\ref{subsubsec:supervised-learning}.

\paragraph{Continual learning from user interactions.}
This can be achieved using reinforcement learning (RL), as described in Section~\ref{subsec:reinforcement-learning}. However, current RL techniques still suffer from requiring a large amount of user-system interaction data, which may have too high a cost in real-world settings. 
To improve the sample efficiency in exploration, three main strategies are considered in the research community \cite{gao2019neural}. 
The first is to encourage the bot to explore dialog states which are less frequently visited, and thus cause more uncertainty for the bot in deciding how to respond \cite{lipton2018bbq,singh2010intrinsically,gasic2010gaussian,pathak2017curiosity}. The second is to use an environment model for efficient exploration, known as model-based RL. In task-oriented dialog, the environment model is a user simulator, which can be used alone (Figure~\ref{fig:deep-dyna}(Middle)) or 
together with human users (Figure~\ref{fig:deep-dyna}(Right)) for dialog policy learning. 
The third strategy is decomposing dialog tasks into smaller subtasks or skills that are easier to learn \cite{peng2017composite,budzianowski2017sub,casanueva2018feudal,tang2018subgoal}. Ideally, we want to identify a set of basic skills that are common across many tasks. After learning these basic skills, a bot can learn to complete more complex tasks more easily via hierarchical reinforcement learning \cite{barto2003recent}.

\paragraph{Updating knowledge.}
The knowledge base does not always contain all information required for task completion. \cite{shen2016implicit} present a knowledge base question-answering bot. The bot is equipped with a ``shared memory'' to store a compact version of the knowledge base. During model training, whenever the bot fails to answer a question because some related information is missing in the knowledge base, the shared memory is updated to incorporate new information. Such a shared memory can be viewed as a long-term memory of the bot that stores knowledge continuously learned from experiences.

\section{A Summary of Preliminary Results}
\label{sec:result}


Evaluating task bots is a challenging research problem \cite{gao2019neural}. 
We group the evaluation methods into two categories: corpus-based evaluation and interactive evaluation.

Corpus-based evaluation is widely used in the research community because it is easy to perform and the evaluation results are reproducible. 
The evaluation relies on a pre-collected, labeled, conversational dataset, where each dialog session is labeled by a user goal, and each dialog turn consists of a dialog history (previous dialog turns) and a ground-truth response, often labeled by a dialog belief state, a database state and a system action presented in dialog acts, e.g., \cite{budzianowski2018multiwoz}. 
Given a user goal and a dialog history, the task bot to be evaluated (target bot) is asked to produce a response. The quality of the target bot is measured by whether the bot-generated responses are fluent and can successfully complete the task according to the user goal.  Corpus-based evaluation does not require the bot to converse with users, but converts an interactive dialog task to a set of one-step response generation tasks for evaluation. 
Thus, the setting is an approximation to how the bot is used in real-world scenarios, where many important capabilities, such as error recovering, cannot be validated. Therefore, corpus-based evaluation is often used to monitor the day-to-day progress of bot development, whereas interactive evaluation with human users has to be performed to verify the quality of the bot before its deployment. 

Interactive evaluation allows users (or user simulators) to converse with the target bot to complete tasks according to a pre-set user goal, and measures the quality of the bot in task success, response appropriateness, and so on \cite{li2020results}. Interactive evaluation is considered more reliable than corpus-based evaluation as the setting of the former closely resembles what the target bot is used in real-world applications. But the interactive evaluation is expensive to perform and the results are not easy to reproduce.

In what follows, we use examples to illustrate how corpus-based evaluation and interactive evaluation are performed, and summarizes the preliminary evaluation result of the GTG-based task bots originally reported in \cite{lei2018sequicity,zhang2019task,Ham2020e2e,hosseini2020simple,peng2020soloist,li2020results}.

\subsection{Corpus-Based Evaluation using MultiWOZ}
\label{subsec:corpus-based-eval}

The Multi-Domain Wizard-of-Oz dataset (MultiWOZ) \cite{budzianowski2018multiwoz} is a fully-labeled collection of human-human written conversations spanning over multiple domains and topics. It contains 10K dialog sessions. Among them 8438, 1000 and 1000 dialog sessions are used for training, validation and testing, respectively. 
Each dialog session contains 1 to 3 domains, including Attraction, Hotel, Hospital, Police, Restaurant, Train, and Taxi. MultiWOZ is inherently challenging due to its multi-domain setting and diverse language styles.

\begin{figure}[t] 
\centering 
\includegraphics[width=1\linewidth]{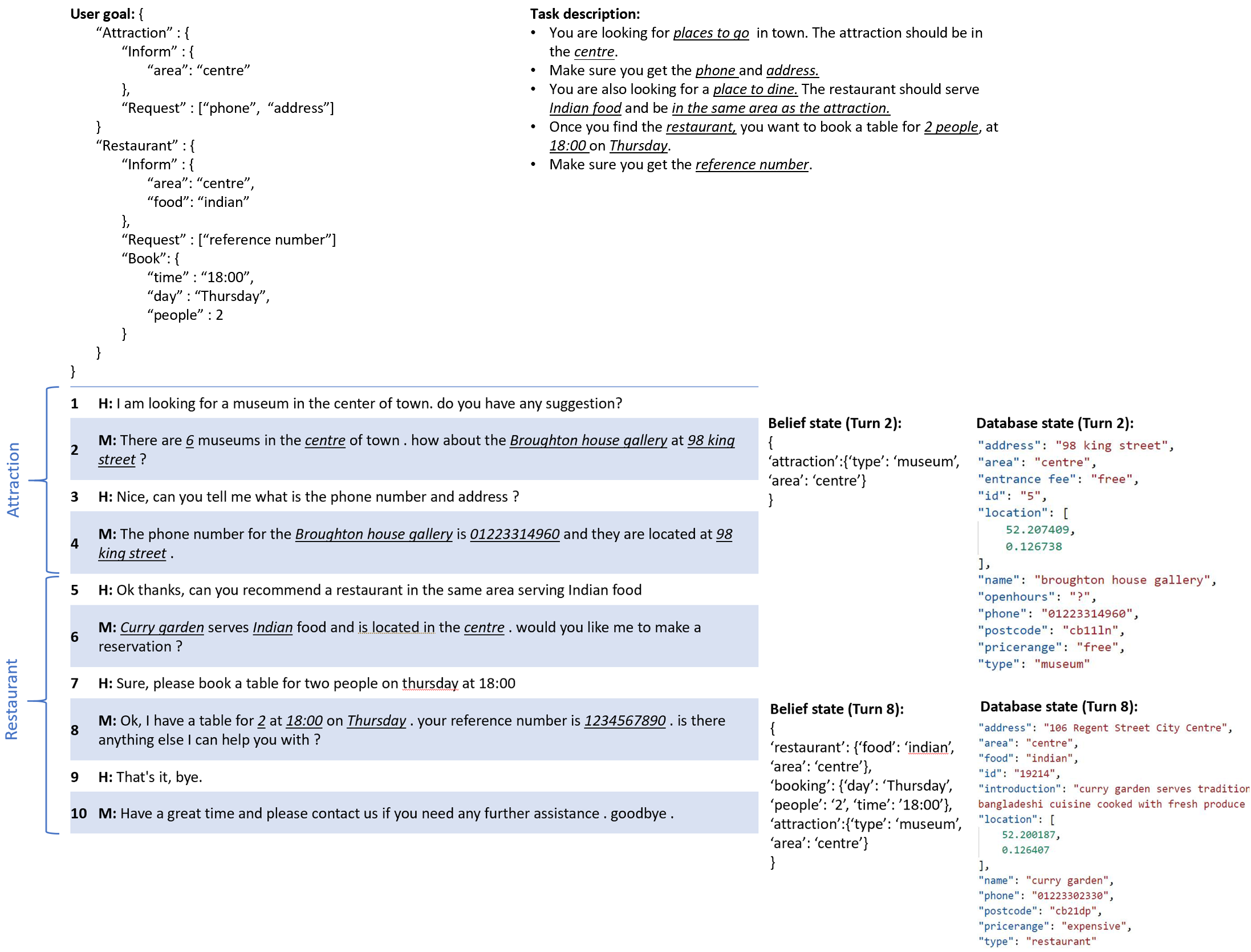}
\vspace{-1mm}
\caption{A multi-domain dialog that consists of (Top) a user goal and a task description, and (Bottom) multiple turns of user-machine utterances, and the dialog belief states and database states at Turns 2 and 8.} 
\label{fig:multiwoz-example} 
\vspace{0mm}
\end{figure}

Figure~\ref{fig:multiwoz-example} shows a multi-domain dialog that consists of a user goal, multiple turns of user-system utterances, and the dialog belief states and database states at Turns 2 and 8.
A user goal consists of three parts. The \emph{informable} slots contain a number of slot-value pairs that serve as constraints the user wants to impose on the dialog. The \emph{requestable} slots are slots whose values are initially unknown to the user and will be filled out during the conversation. The \emph{book} slots are used to reserve a place recommended by the system. When the system offers to book and the user accepts it, the system should notify an 8-digit reference number.
The dialog belief state is the bot's internal cognitive model that stores information about the user goal distilled from the user-system conversation. The database state contains the entities retrieved from the task-specific database using the belief state as a query. 

Budzianowski et al. \cite{budzianowski2018multiwoz} recommend to break down dialog modeling into three sub-tasks and report a benchmark result for each using the MultiWOZ dataset: dialog state tracking, dialog-act-to-text generation (or system action selection using dialog policy), and dialog-context-to-text generation (or natural language generation). 
While these sub-tasks are useful for component-wise evaluation, we advocate to perform an end-to-end, system-wise evaluation using MultiWOZ for two reasons. 
The first reason is that the breakdown is based on the classical modular architecture of task bots, as shown in Figure~\ref{fig:two-dialogue-system}(Top). However, many recently proposed task bots do not follow the modular architecture, but use joint-models \cite{MDRG,chen2019semantically,zhao2019rethinking} or even an end-to-end system architecture \cite{lei2018sequicity,zhang2019task}.  The second reason is that the component-wise evaluation results are not always consistent with the overall performance of dialog systems \cite{takanobu2020your}.

\begin{figure}[t] 
\centering 
\includegraphics[width=1\linewidth]{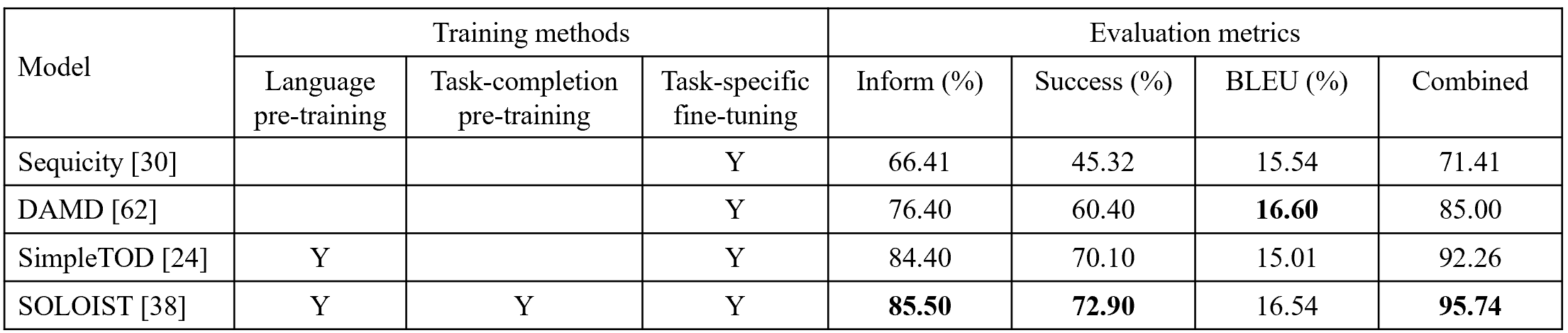}
\vspace{-3mm}
\caption{Corpus-based end-to-end evaluation results on MultiWOZ.} 
\label{fig:eval-multiwoz} 
\vspace{0mm}
\end{figure}

Figure~\ref{fig:eval-multiwoz} shows the system-wise evaluation results of four dialog systems on MultiWOZ, originally reported in \cite{lei2018sequicity,zhang2019task,hosseini2020simple,peng2020soloist}. The qualities of these task bots are measured using three metrics: Inform, Success and  BLEU \cite{papineni2002bleu}. The first two metrics relate to the dialogue task completion - whether the system has provided an appropriate entity (Inform) and then answered all the requested attributes (Success). The BLEU score measures how fluent the generated responses are. A combined score (Combined) is also reported using 
Combined = (Inform + Success) $\times$ 0.5 + BLEU
as an overall quality measure \cite{budzianowski2018multiwoz}. 

The four systems in Table 1 are developed using hybrid approaches similar to what is described in Section \ref{subsec:gtg4dialog}, in that they all follow the classical modular architecture of Figure~\ref{fig:two-dialogue-system}(Top) for system design but use neural networks for system implementation. They differ in the neural network architecture employed and the way the dialog models are trained.

Sequicity \cite{lei2018sequicity} casts dialog state tracking as a task of detecting belief span which is a text span that tracks the belief states at each turn, and then decomposes the task-oriented dialog pipeline (Figure~\ref{fig:two-dialogue-system}(Top)) into three generation tasks, performed in three stages, respectively.
In the first stage, Sequicity generates belief spans from dialog history to facilitate KB search, as GTG's state tracking and knowledge based lookup of Equations \ref{eq:dst} and \ref{eq:lookup}, respectively. 
In the second stage, it generates a system action grounded in dialog belief state and KB search results, as in Equation \ref{eq:gra}.  Sequicity employs a seq2seq model, dubbed Two-Stage CopyNet (TSCP), which consists of a sequence of two RNNs to handle the two aforementioned generation tasks, respectively. The two RNNs are trained separately on labeled task-specific dialog datasets. In the third stage, a separate NLG module is used to generate the natural language response using the generated system action as in Equation \ref{eq:nlg}.

The Domain Aware Multi-Decoder (DAMD) network \cite{zhang2019task} uses the same model architecture as Sequicity (TSCP), but is managed to get better results than Sequicity on multi-domain task-oriented dialog by training the neural dialog models on a more diverse and larger conversational dataset augmented using the proposed Multi-Action Data Augmentation (MADA) method. MADA discovers from a pre-collected conversational dataset the mapping from one dialog state (condensed dialog history) to multiple system actions. Then during model training, DAMD is learned to generate, from dialog state, not only its ground-truth system action in the original training data, but also the system actions discovered by MADA.

SOLOSIT \cite{peng2020soloist} is the GTG-based task bot described in Section \ref{subsec:gtg4dialog}. GTG is pre-trained and fine-tuned for the completion of specific tasks in three stages: (1) language pre-training (through the use of GPT-2), (2) task completion pre-training, and (3) task-specific fine-tuning, as described in Section \ref{sec:introduction}.

SimpleTOD \cite{hosseini2020simple} is similar to SOLOIST in that both cast the classical modular dialog system as a sequential data generation pipeline and implement the pipeline using a Transformer-based model, which is initialized using a pre-trained language model (GPT-2) and fine-tuned on task-specific data.  
However, SimpleTOD differs from SOLOIST in some important implementation details. SimpleTOD treats POL and NLG as two separate steps, and requires more expensive annotations (i.e., system actions in the form of dialog acts) for training.  
As a results, SimpleTOD cannot effectively use heterogeneous dialog data for task completion pre-training as discussed in Section \ref{subsec:gtg4dialog}, and thus is trained in two stages: (1) language pre-training (through the use of GPT-2), and (2) task-specific fine-tuning on the pre-trained language model.

The results in Figure~\ref{fig:eval-multiwoz} show that although all these systems are developed using similar hybrid approaches (i.e., using the classical modular system architecture but implemented using neural network models), their performances depend to a large degree upon how effective they can leverage training data. 
Sequicity and DAMD use only task-specific dialog dataset for model training. 
DAMD outperforms Sequicity mainly because DAMD uses more training data generated by the proposed data augmentation method. 
SimpleTOD uses the GPT-2 model, which is pre-trained on large amounts of raw text, as an initial model to adapt to individual dialog tasks using task-specific data. Thus, it significantly outperforms Sequicity and DAMD which do not use GPT-2. 
SOLOIST is the best performer mainly because it uses the heterogeneous dialog datasets to pre-train the GTG model, initialized by GPT-2, and then fine-tune GTG to individual tasks using task-specific data.

As detailed in \cite{peng2020soloist}, SOLOIST can be easily adapted to a new task in the few-shot learning setting, where for each new task less than 50 task-specific dialog sessions are available for fine-tuning. This is a more realistic setting than the standard MultiWOZ setting where hundreds to thousands of dialogs exist for each task. 
SOLOIST is reported to outperform other systems more significantly in few-shot learning than in the standard MultiWOZ setting (Figure~\ref{fig:eval-multiwoz}), which attributes to a large degree to the use of heterogeneous dialog data for task-completion pre-training. 
\cite{peng2020soloist} also report that using the machine teaching tool of conversation learner significantly improves the effectiveness of fine-tuning, substantially lowering the labeling cost.
However, widely-accepted research benchmarks for evaluating few-shot learning and machine teaching for task bots are yet to be developed to enable a more comprehensive study.  

\subsection{Interactive Evaluation using ConvLab}
\label{subsec:interactive-eval}

ConvLab \cite{lee2019convlab,zhu-etal-2020-convlab} is an open-source platform that supports researchers to train and evaluate their own dialog systems.
As shown Figure~\ref{fig:convlab2},
to support interactive evaluation, ConvLab provides a user simulator for automatic evaluation and integrates Amazon Mechanical Turk for human evaluation.

The user simulator is agenda-based \cite{schatzmann2007agenda}. 
It consists of a multi-intent language understanding (MILU) \cite{lee2019sumbt,hakkani2016multi} for NLU, a rule-based policy, and a template-based NLG.
For each dialogue, a user goal is randomly generated that conforms with the goal schema of MultiWOZ (Figure~\ref{fig:multiwoz-example}(Top)). 
Then, the simulator’s policy uses a stack-like agenda with complex hand-crafted heuristics to inform its goal and mimics complex user behaviors during a conversation. Since the system interacts with the simulator in natural language, the user simulator directly takes system utterances as input and outputs a user response.
The simulator is an approximation to human users. 
Although automatic evaluation using simulators shows a moderate correlation with human evaluation, it could remarkably overestimate the system performance in human interactions \cite{li2020results,takanobu2020your}.

\begin{figure}[t] 
\centering 
\includegraphics[width=1\linewidth]{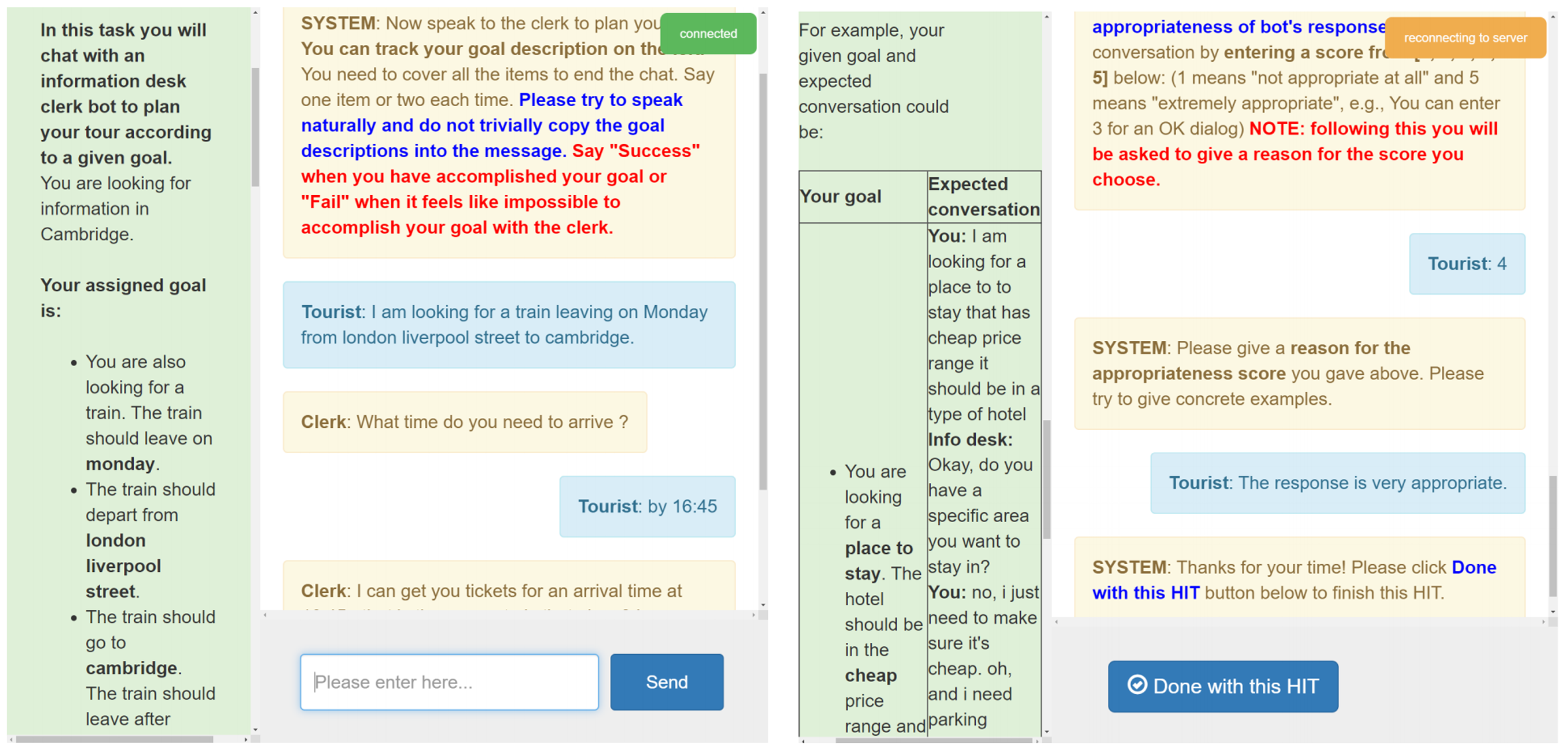}
\vspace{-1mm}
\caption{Human interactive evaluation using MTurk \cite{li2020results}. (Left) A human-system conversation starts. (Right) The conversation ends.} 
\label{fig:human-eval} 
\vspace{0mm}
\end{figure}

For the human evaluation, the target dialog system is hosted
in ConvLab as a bot service and the human evaluation is crowdsourced on Amazon Mechanical Turk as illustrated in Figure~\ref{fig:human-eval}.
In each conversation, the system samples a goal and presents it to the MTurker with instructions. Then the MTurker converses
with the system via natural language to achieve the goal and
judges the system based on three metrics: 
(1) Success/Failure evaluates task completion; 
(2) Language Understanding Score (ranging from 1 (bad) to 5 (good)) indicates the extend to which the bot understands user inputs; and
(3) Response Appropriateness Score (ranging from 1 (bad) to 5 (good)) measures whether the response is appropriate and fluent. 

\begin{figure}[t] 
\centering 
\includegraphics[width=1\linewidth]{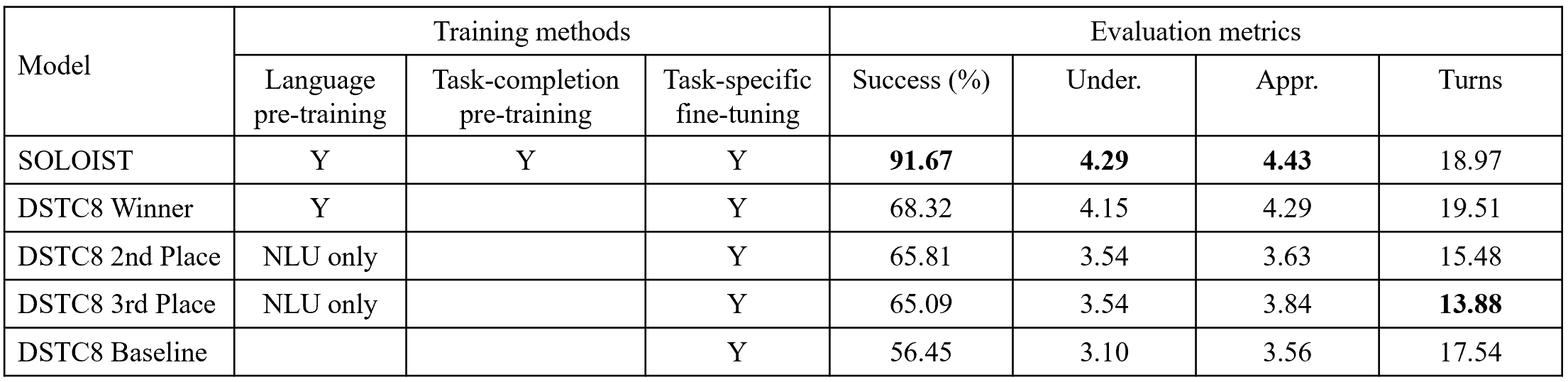}
\vspace{-3mm}
\caption{Human interactive evaluation results using the setting of Track 1 at DSTC-8 \cite{li2020results,peng2020soloist}.} 
\label{fig:eval-dstc8} 
\vspace{0mm}
\end{figure}

Figure~\ref{fig:eval-dstc8} shows the human interactive evaluation results using the setting of the ``multi-domain task completion'' track (Track 1) at the 8th Dialog System Technology Challenge (DSTC-8), originally reported in \cite{li2020results,peng2020soloist}.
The results are consistent with the corpus-based evaluation results in Figure~\ref{fig:eval-multiwoz}. The performances of these systems depend to a large degree upon how effective they can leverage training data of different types.
SOLOIST is trained in three stages: language pre-training on raw text, task completion pre-training on heterogeneous dialog data, and fine-tuning on task-specific data. 
The DSTC-8 Winner system is based on the end-to-end neural pipeline model \cite{Ham2020e2e}. Like SimpleTOD, the neural pipeline model needs the annotations of system actions for training, and is directly fine-tuned from the pre-trained GPT-2 model to individual dialog tasks using task-specific labeled data. Unlike SOLOIST, the neural pipeline model is not pre-trained for task-completion using heterogeneous dialog data.   
The other systems are the classical modular systems, composed of machine-learned and rule-based modules. 
The two systems, which are ranked the second and third places at DSTC-8 respectively, use BERT-based NLU modules that are fine-tuned on task-specific data, and rule-based modules for DST, POL and NLG. 
The DSTC-8 baseline is composed of dialog modules implemented in ConvLab, including a MILU model trained on task-specific data for NLU, a rule-based DST, a rule-based dialog policy, and a template-based NLG module.

\begin{figure}[t] 
\centering 
\includegraphics[width=1\linewidth]{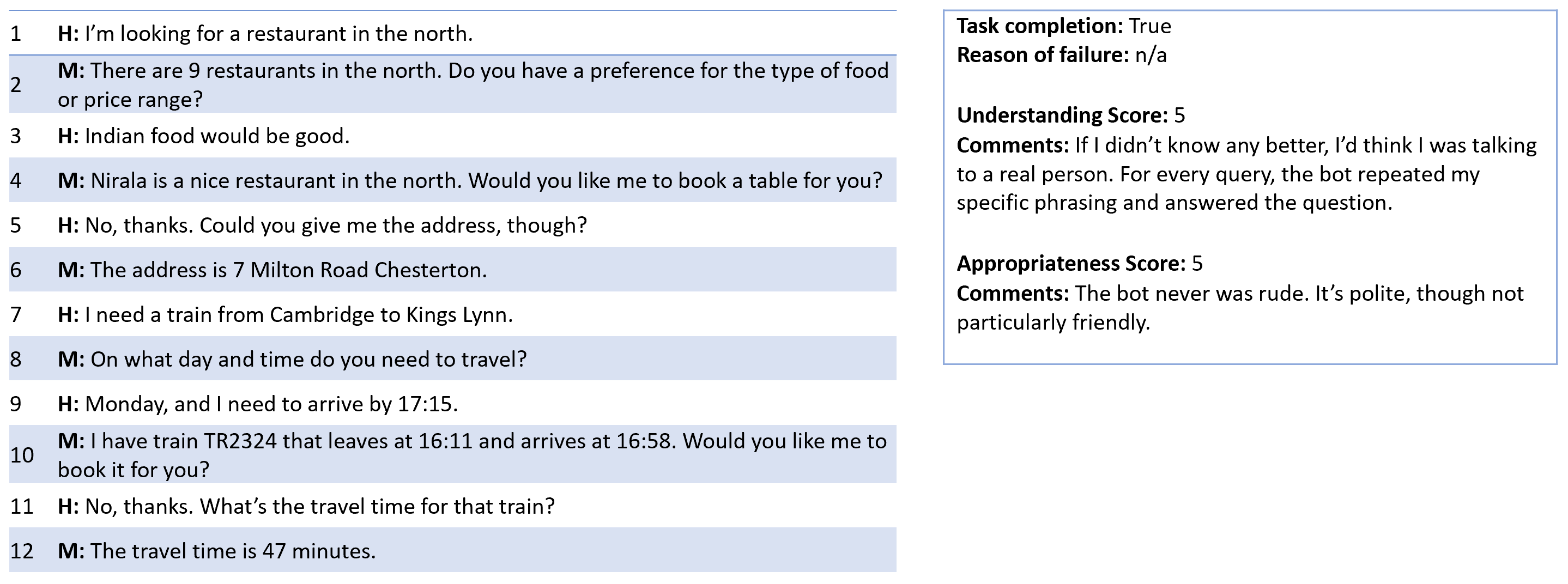}
\vspace{-2mm}
\caption{A dialog session generated by MTurks and SOLOIST, used for interactive evaluation. Human judges and comments are attached. The bot generates fluent responses and completes the task successfully.} 
\label{fig:interactive-eval-example-1} 
\vspace{0mm}
\end{figure}

\begin{figure}[t] 
\centering 
\includegraphics[width=1\linewidth]{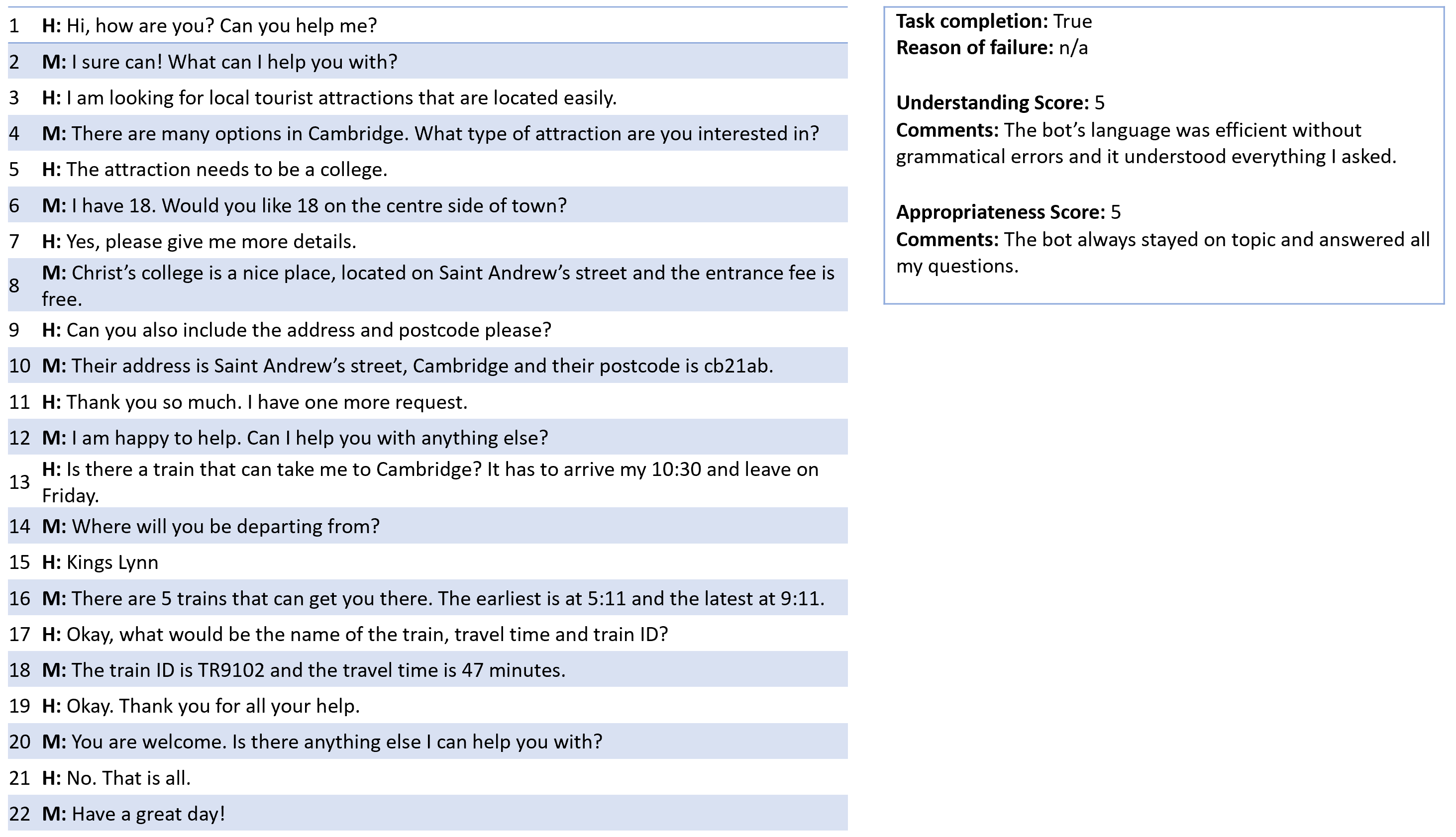}
\vspace{-2mm}
\caption{A dialog session generated by MTurks and SOLOIST, used for interactive evaluation. Human judges and comments are attached. The bot always stays on topic, answers all the user questions, and completes the task successfully.} 
\label{fig:interactive-eval-example-2} 
\vspace{0mm}
\end{figure}

\begin{figure}[t] 
\centering 
\includegraphics[width=1\linewidth]{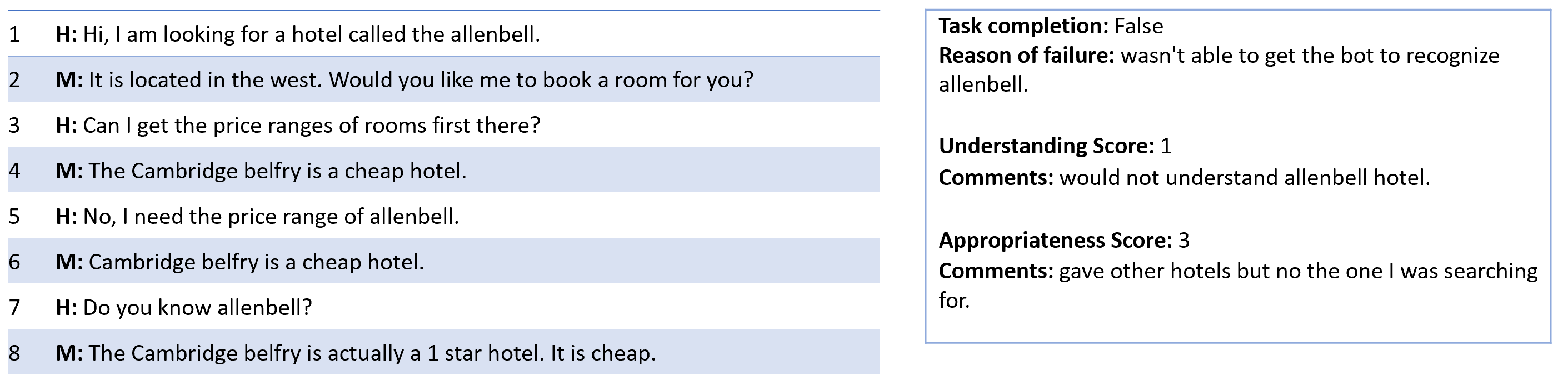}
\vspace{-2mm}
\caption{A dialog session generated by MTurks and SOLOIST, used for interactive evaluation. Human judges and comments are attached. The bot failed to complete the task because it failed to recognize ``allenbell'' as an entity mention.} 
\label{fig:interactive-eval-example-3} 
\vspace{0mm}
\end{figure}

\begin{figure}[t] 
\centering 
\includegraphics[width=1\linewidth]{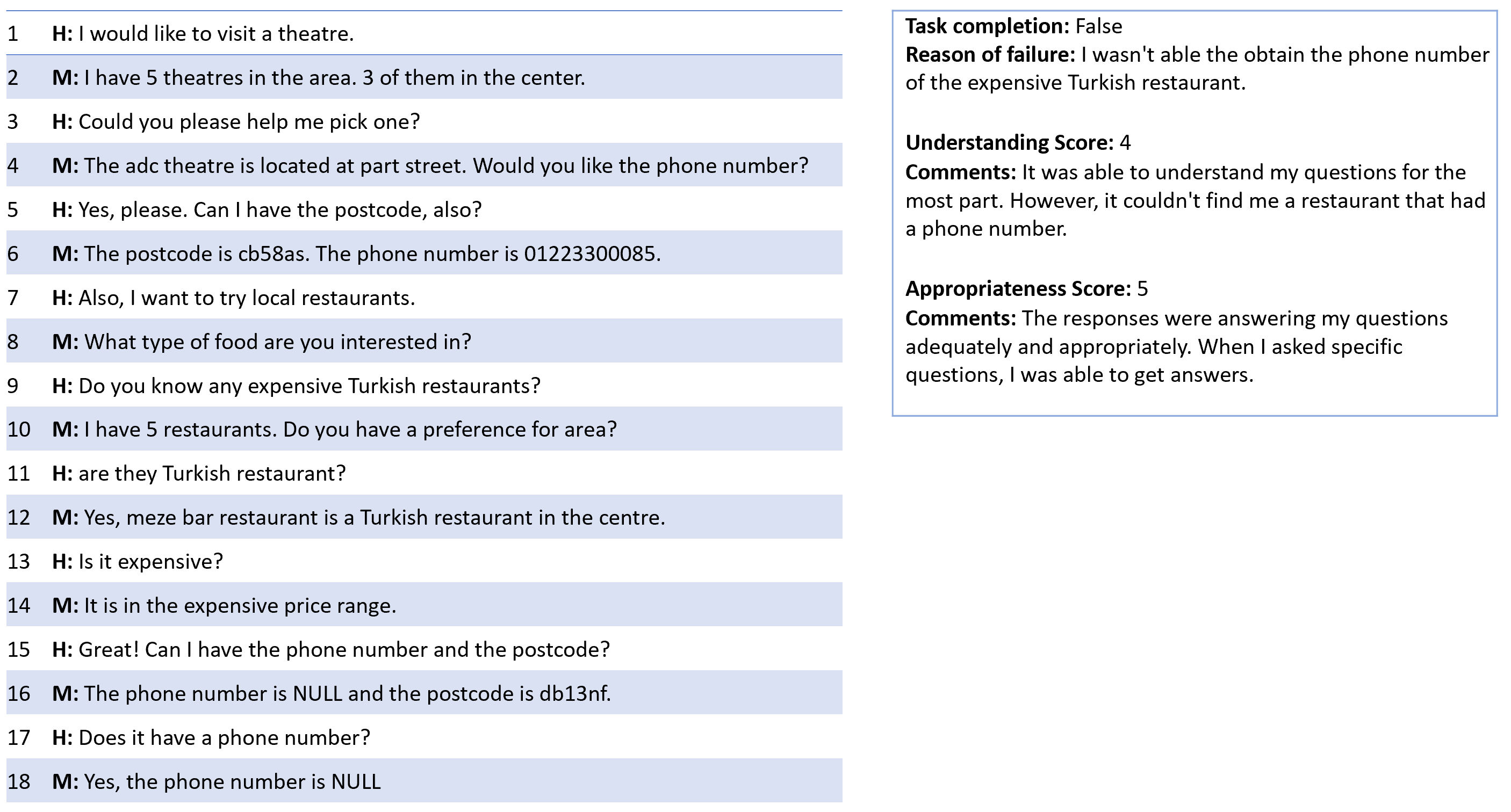}
\vspace{-2mm}
\caption{A dialog session generated by MTurks and SOLOIST, used for interactive evaluation. Human judges and comments are attached. The bot failed to complete the task because the phone number of the Meza Bar Restaurant is missing in the database.} 
\label{fig:interactive-eval-example-4} 
\vspace{0mm}
\end{figure}

We see that SOLOIST outperforms the other systems by much larger margins in Success in human interactive evaluation in Figure~\ref{fig:human-eval} than that in corpus-based evaluation in Figure~\ref{fig:eval-multiwoz}. 
\cite{peng2020soloist} attribute this to the fact that Turks use more diverse language to interact with the target bots in interactive evaluation than that in the pre-collected MultiWOZ dataset and the use of heterogeneous dialog data for task-completion pre-training makes SOLOIST a more robust task bot than the others. 
Figures~\ref{fig:interactive-eval-example-1} to~\ref{fig:interactive-eval-example-4} present four dialog examples generated by SOLOIST and human users (MTurks), and their associated human judges and comments, used in our interactive evaluation. We see that the bot-generated responses are fluent and on topic most of the time so that a human users commented that it was like talking to a real person. The dialog in Figure~\ref{fig:interactive-eval-example-3} is failed because the bot fails to recognize ``allenbell'' as an entity mention. This problem could have been addressed by incorporating a task-specific named entity recognizer as described in Section~\ref{subsec:sym-modules}.
The dialog in Figure~\ref{fig:interactive-eval-example-4} is failed because the phone number of the Meza Bar Restaurant is missing in the database. This problem could have been addressed by updating the database as described in Section~\ref{subsec:continual-learning}.

\section{Discussions and Future Work}
\label{sec:conclusion}

\begin{figure}[t] 
\centering 
\includegraphics[width=0.95\linewidth]{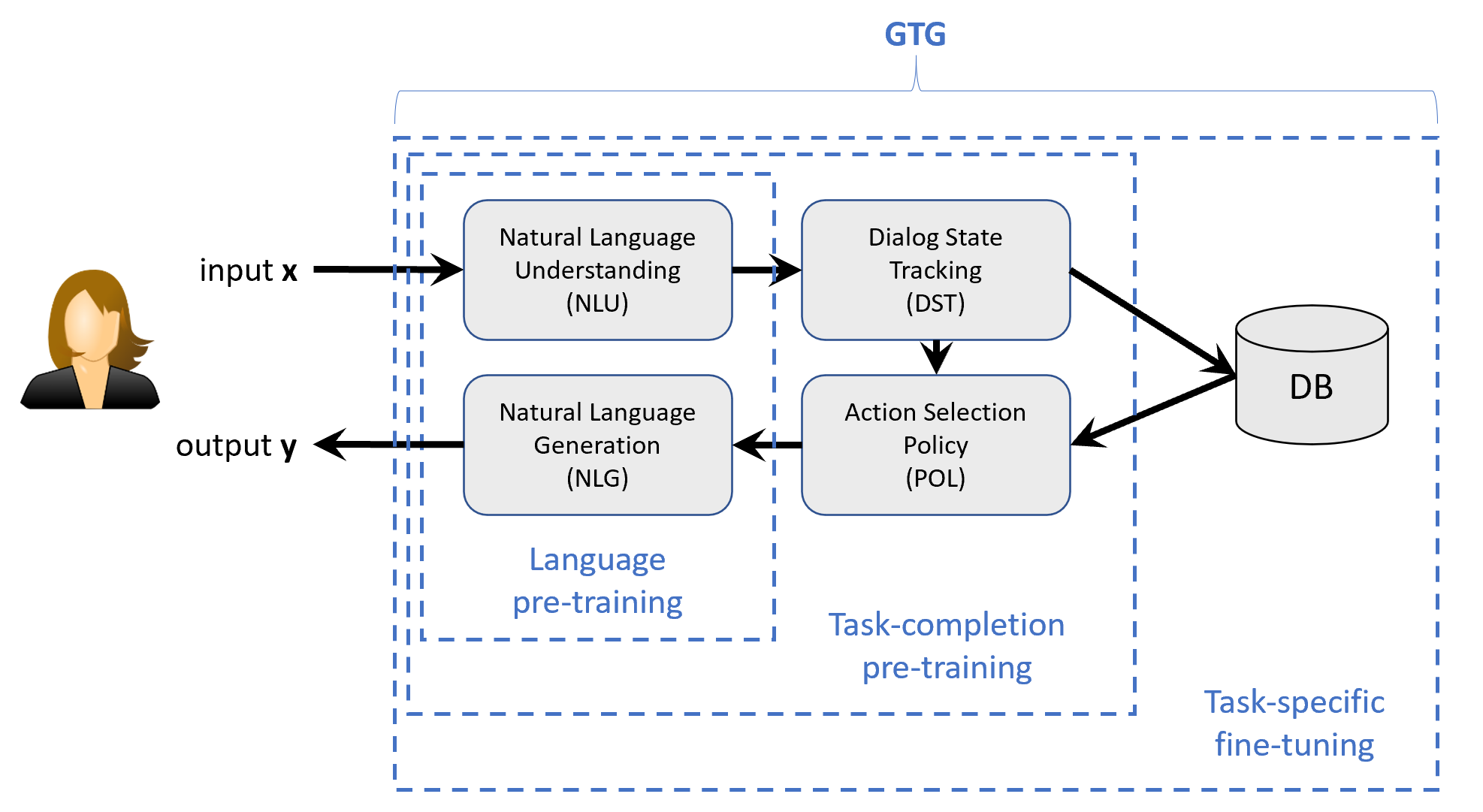}
\vspace{0mm}
\caption{The GTG-based hybrid approach to building task bots at scale. It follows the classical modular architecture of task-oriented dialog systems. The modules are implemented using a hybrid GTG model that employs a multi-layer Transformer neural network as the backbone, combined with symbol-manipulation modules for knowledge base inference and prior knowledge encoding. GTG is trained in three stages for task completion: (1) language pre-training on raw text, (2) task completion pre-training on heterogeneous dialog data, (3) task-specific fine-tuning on task-specific training dialogs.} 
\label{fig:gtg-training} 
\vspace{0mm}
\end{figure}

This paper presents a GTG-based hybrid approach to building robust task bots at scale.  
As illustrated in Figure~\ref{fig:gtg-training}, our approach follows the classical modular architecture of task-oriented dialog systems to make the behavior of a bot reliable, interpretable and controllable. The bot is equipped with a cognitive model to keep track of the dialog state and generates responses grounded in dialog state and task-specific knowledge for task completion. Our approach follows the pre-training and fine-tuning framework of transfer learning to build bots at scale. The dialog modules are implemented using a hybrid GTG model that employs a multi-layer Transformer neural network as the backbone, combined with symbol-manipulation modules for knowledge base inference and prior knowledge encoding. GTG is trained in three stages for task completion: (1) language pre-training on raw text, (2) task completion pre-training on heterogeneous dialog data, (3) task-specific fine-tuning on task-specific training dialogs.
The three-stage learning allows task bots to transfer the knowledge and skills learned in one task to others. For example, the user-bot dialogs logged after the deployment of the bots can be used to improve the GTG pre-training and consequently improves all the (future) bots that are fine-tuned on GTG. 

Although we have witnessed promising results of the approach in both corpus-based evaluation and interactive evaluation on research benchmarks (Section 5), several problems need to be addressed to apply the approach for building real-world task bots at scale. We brief some of most urgent ones below.
As pointed out in Section \ref{subsec:continual-learning}, the capability of continual learning is crucial for a task bot to continuously improve its performance and adapt itself to the dynamic environment after its deployment. We envision that continual learning is a human-in-the-loop ML process. We are improving CL by providing a unified machine teaching UI that allows dialog authors to easily collect dialog logs, correct bot's mistakes, incorporate new rules, update knowledge bases, and so on. 
Using larger pre-trained models (such as GPT-3) is likely to improve all GTG-based bots. However, the high cost of maintaining and hosting large models prevents us from deploying these models for many real-world applications running on resource-limited devices. We are exploring task-agnostic knowledge distillation methods to significantly reduce the model size. 

The GTG-based hybrid approach is just a baby step towards robust conversational AI in that the approach heavily relies on supervised learning where a bot learns mainly by following antecedent experiences, and thus while it can respond reliably to input in an accustomed environment, it is unable to quickly adjust to changes in the environment. 
Sutton and Barto \cite{sutton2018reinforcement} propose to develop a model-based agent which learns to control its behaviors by knowledge of its goals and the relation between actions and their consequences, and thus can quickly accommodate changes in its environment. 
The DDQ agent described in Section~\ref{subsec:reinforcement-learning} and Figure~\ref{fig:deep-dyna}(Right) is an instance of such a model-based agent. The DDQ agent has an environment model (user simulator) consisting of an agenda-based state-transition model and a reward model, and can decide how to respond by using the model to simulate sequences of response choices (planning) to find a path that results in the highest reward.  Therefore, any change in environment, if it is captured by the user simulator (through the environment model learning process), automatically leads to an updated dialog policy via the planning process.  
We believe that a GTG model that can be learned via model-based RL such as DDQ lays the foundation of more promising approaches to robust conversational AI.

It is imperative to make our bot building approach more scalable by improving compositionality generation -- the ability to generalize to novel compositions of familiar constituents. In the dialog context, these familiar constituents are primary dialog skills that are highly reusable and can be composed to deal with various, more sophisticated, dialog tasks. Some researchers have begun scratching the surface of it by applying hierarchical RL or feudal RL for composite tasks, each consisting of a set of subtasks that need to be collectively solved using primary dialog skills \cite{peng2017composite, cuayahuitl2010evaluation, budzianowski2017sub, casanueva2018feudal, tang2018subgoal}. Moreover, there are studies on learning to compose primary skills to form a wide variety of complex skills which can be unseen compositions, allowing for zero-shot generalization \cite{sahni2017learning,andreas2016learning}.

The GTG model described in this paper is developed for task bots. We like to expend it to build social bots for various social interaction tasks ranging from persuasion \cite{wang-etal-2019-persuasion,tan2016winning} to AI companion \cite{zhou2020design}. As pointed out in \cite{zhou2020design}, such a social bot needs to demonstrate a consistent personality, and has a rich set of IQ and EQ skills to generate interpersonal responses to establish long-term emotional connections with users. So, the bot needs to be grounded in a concrete social context: role, status, intention, region, ethnicity, social networks, and so on \cite{wardhaugh2011introduction}.  Since these types of social grounding information and signals are unlikely labeled in any pre-collected corpora, it is critical to construct an interactive learning setting similar to the RL setting in Section~\ref{subsec:reinforcement-learning}, where the social bot can be trained via social interactions with users in natural situations and social communities. 

\bibliographystyle{plain}
\bibliography{references}
\end{document}